\documentclass[10pt,twocolumn,letterpaper]{article}

\usepackage{cvpr}              
\definecolor{cvprblue}{rgb}{0.21,0.49,0.74}
\usepackage[pagebackref,breaklinks,colorlinks,allcolors=cvprblue]{hyperref}
\usepackage{booktabs,makecell, multirow, tabularx}
\usepackage{stfloats}

\title{VGGDrive: Empowering Vision-Language Models with Cross-View Geometric Grounding for Autonomous Driving}

\author{
Jie Wang\textsuperscript{1,2\dag}\quad
Guang Li\textsuperscript{2}\quad
Zhijian Huang\textsuperscript{2}\quad \\
Chenxu Dang\textsuperscript{2}\quad 
Hangjun Ye\textsuperscript{2}\quad
Yahong Han\textsuperscript{1*}\quad
Long Chen\textsuperscript{2}
\\[0.3cm]
\textsuperscript{1}College of Intelligence and Computing, Tianjin University \quad
\textsuperscript{2}Xiaomi EV \\ \url{https://github.com/WJ-CV/VGGDrive}
}

\begin{document}
\maketitle
\footnotetext{
    Work done while interning at Xiaomi Embodied Intelligence Team.\\
    \makebox[1.2em]{} *Corresponding author. \texttt{yahong@tju.edu.cn} \\ 
    \makebox[1.6em]{}Primary contact: Jie Wang. \texttt{wangjiexy@tju.edu.cn}
}
\begin{abstract}
The significance of cross-view 3D geometric modeling capabilities for autonomous driving is self-evident, yet existing Vision-Language Models (VLMs) inherently lack this capability, resulting in their mediocre performance. While some promising approaches attempt to mitigate this by constructing Q$\&$A data for auxiliary training, they still fail to fundamentally equip VLMs with the ability to comprehensively handle diverse evaluation protocols. We thus chart a new course, advocating for the infusion of VLMs with the cross-view geometric grounding of mature 3D foundation models, closing this critical capability gap in autonomous driving. In this spirit, we propose a novel architecture, \textbf{VGGDrive}, which empowers \textbf{V}ision-language models with cross-view \textbf{G}eometric \textbf{G}rounding for autonomous \textbf{Driv}ing. Concretely, to bridge the cross-view 3D geometric features from the frozen visual 3D model with the VLM's 2D visual features, we introduce a plug-and-play Cross-View 3D Geometric Enabler (CVGE). The CVGE decouples the base VLM architecture and effectively empowers the VLM with 3D features through a hierarchical adaptive injection mechanism. Extensive experiments show that VGGDrive enhances base VLM performance across five autonomous driving benchmarks, including tasks like cross-view risk perception, motion prediction, and trajectory planning. It’s our belief that mature 3D foundation models can empower autonomous driving tasks through effective integration, and we hope our initial exploration demonstrates the potential of this paradigm to the autonomous driving community.
\end{abstract}
\vspace{-0.2in}

\section{Introduction}
\label{sec:intro}

\begin{figure}[!t]
	\centering
	\includegraphics[width=3.26in]{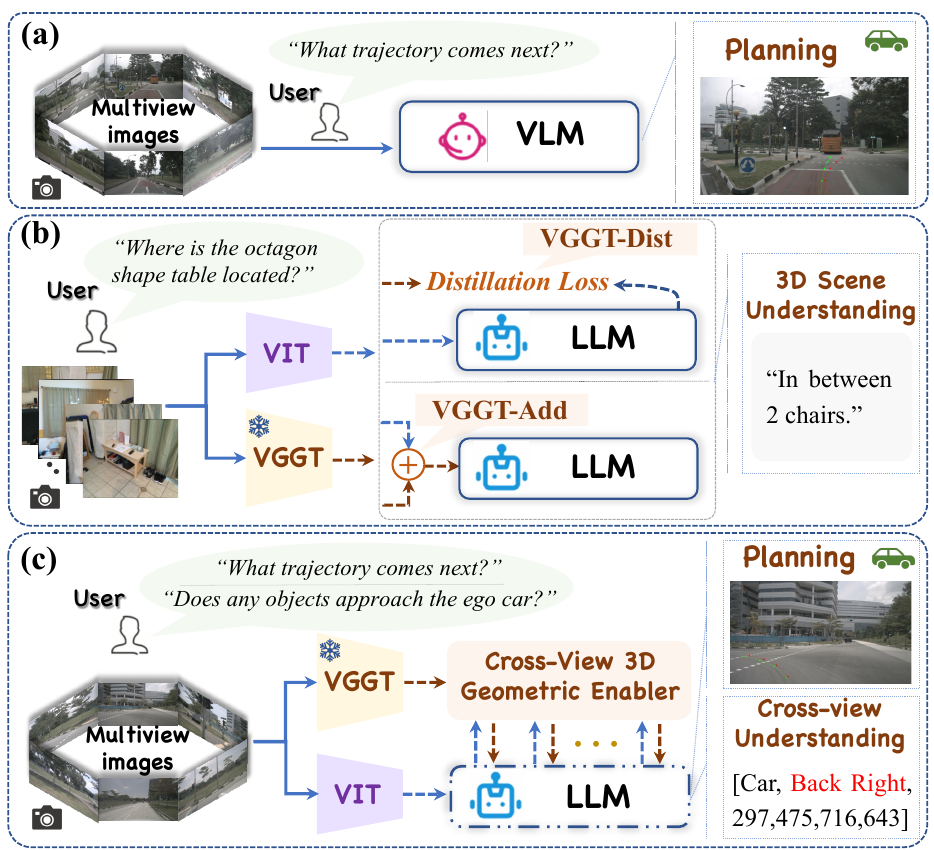}
	\vspace{-0.25in}
	\DeclareGraphicsExtensions.
	\begin{center}
		\caption{Existing relevant paradigms \textit{vs.} our VGGDrive. (a) The VLA paradigm for trajectory planning. (b) Two existing paradigms for integrating 3D foundation models (VGGT \cite{wang2025vggt}) with VLMs: VGGT-Dist \cite{huang2025mllms} and VGGT-Add \cite{zheng2025learning}. (c) Our VGGDrive, which leverages the VGGT model to profoundly empower the basic VLM with cross-view geometric grounding capabilities, thereby handling diverse autonomous driving tasks.} \label{fig-1}
	\end{center}
	\vspace{-0.4in}
\end{figure}

Visual-Language Models (VLMs) \cite{zhu2025internvl3, bai2025qwen2, wang2024qwen2}, equipped with extensive world knowledge and powerful reasoning capabilities acquired from vast internet-scale datasets, offer a promising pathway to overcome the generalization bottleneck faced by traditional autonomous driving systems \cite{hu2023planning, jiang2023vad, weng2024drive}. Early studies \cite{jiang2024senna, jiang2025alphadrive, marcu2024lingoqa} primarily employed VLMs as high-level scene understanding and decision-making aids, generating natural language descriptions of the environment or driving suggestions by analyzing visual inputs. With the maturation of end-to-end learning paradigms \cite{hwang2024emma, renz2025simlingo, li2025recogdrive, luo2025adathinkdrive, huang2024making}, the research focus has evolved toward the more promising Vision-Language-Action (VLA) models (Fig.\ref{fig-1} (a)). Within this unified VLM framework, configured through diverse training data and task-specific instructions, VLMs exhibit significant task adaptability: they \cite{renz2024carllava, renz2025simlingo} can perform deep scene perception and reasoning based on their semantic priors, and directly translate semantic understanding into concrete driving trajectories.

\begin{figure}[!t]
	\centering
	\includegraphics[width=3.26in]{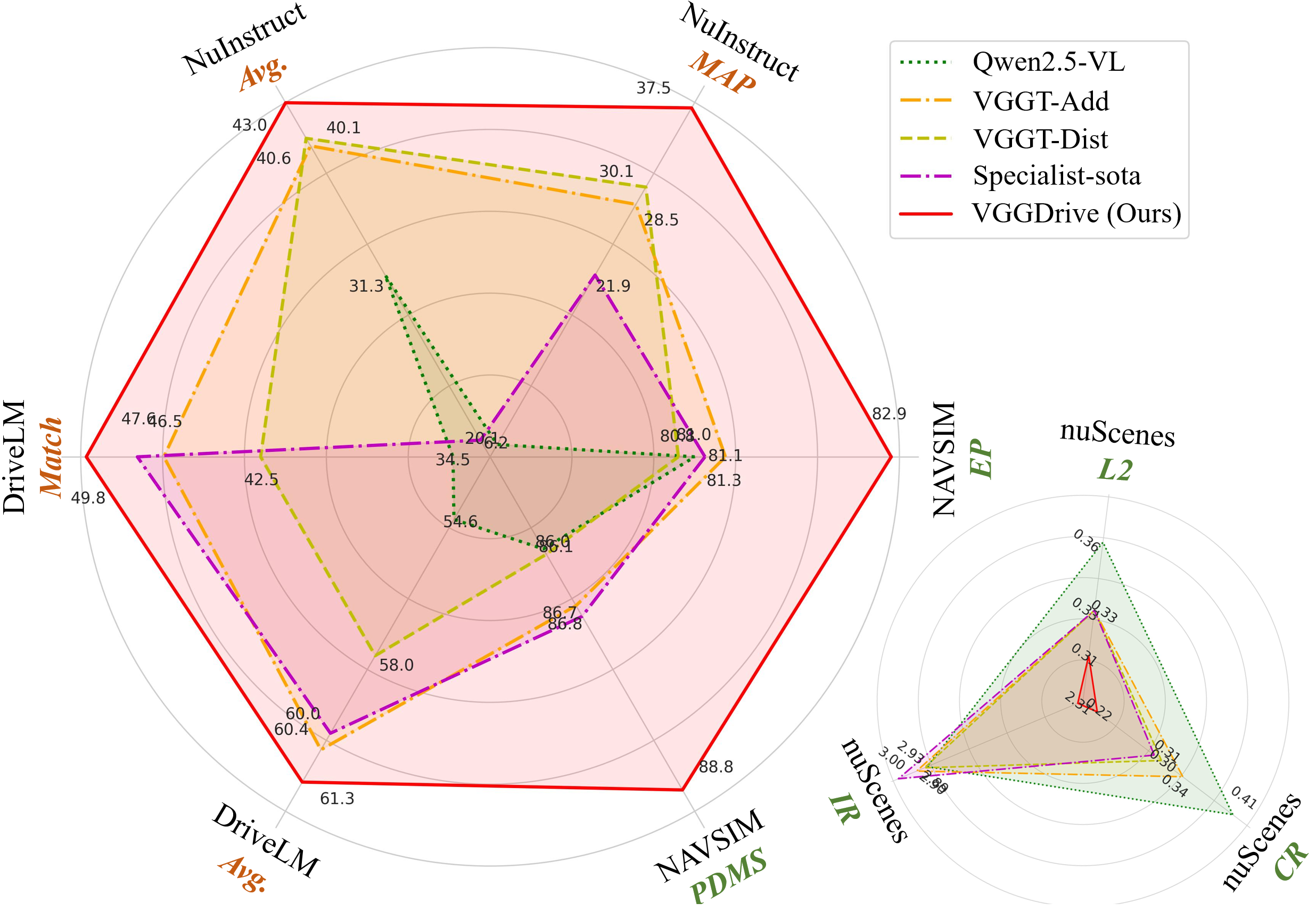}
	\vspace{-0.2in}
	\DeclareGraphicsExtensions.
	\begin{center}
		\caption{Quantitative comparison of VGGDrive with specific sota methods across four autonomous driving benchmarks, covering evaluations of attributes such as cross-view risk \textit{perception}, motion \textit{prediction} and trajectory \textit{planning}.} \label{fig-2}
	\end{center}
	\vspace{-0.4in}
\end{figure}

However, the practical performance of this paradigm \cite{renz2024carllava, fu2025orion, hwang2024emma} remains constrained by a critical bottleneck: safe navigation in complex, open environments for autonomous driving heavily relies on precise spatial perception capabilities, while VLMs inherently lack the ability to model cross-view geometry in the 3D physical world. This fundamental limitation hampers the model’s performance in real-world autonomous driving tasks (e.g., Qwen2.5-VL \cite{bai2025qwen2} in Fig. \ref{fig-2}), where fine-grained spatial understanding is essential.

To mitigate this gap, several studies \cite{wang2025omnidrive, sima2024drivelm, ding2024holistic, gholami2025spatial} have attempted to teach VLMs spatial concepts by constructing large-scale, task-specific question-answer (Q$\&$A) datasets. However, these approaches struggle to fundamentally equip the model with solid geometric priors, yielding only limited improvements (Specialist-sota in Fig. \ref{fig-2}). Consequently, subsequent works \cite{li2025recogdrive, luo2025adathinkdrive} have resorted to a ``compromise" technical route: introducing an independent action decoder on top of the VLM to specialize in trajectory prediction. While this approach improves trajectory performance, it disconnects scene understanding from decision-making, preventing the model's knowledge and reasoning from being effectively translated into final control outputs.



Concurrently, powerful visual 3D foundation models \cite{wang2025vggt, leroy2024grounding, wang2024dust3r} have emerged from pre-training on massive 3D datasets, demonstrating superior cross-view geometric modeling. Several pioneering works \cite{huang2025mllms, zheng2025learning, abouzeid2025geoaware, li2025towards, lin2025evo, li2025spatial} have shown that integrating VLMs with VGGT \cite{wang2025vggt} holds significant potential for tasks such as indoor 3D scene understanding. However, these methods \cite{huang2025mllms, zheng2025learning, abouzeid2025geoaware} are predominantly developed using indoor, static, monocular video sequences, which significantly differ from the outdoor, dynamic, multi-camera complex environments encountered in autonomous driving. Furthermore, existing methods implement relatively simple integration schemes (as shown in Fig. \ref{fig-1}(b)), which fail to effectively empower VLMs to meet the high precision and robustness requirements of complex autonomous driving scenarios (as detailed in Fig. \ref{fig-2}).

At this juncture, we ponder: \textit{How can VGGT \cite{wang2025vggt} effectively empower VLMs with cross-view geometric grounding to compensate for their inherent limitations and advance autonomous driving capabilities?}

In this spirit, we propose a novel architecture: \textbf{VGGDrive}, empowering \textbf{V}ision-language models with cross-view \textbf{G}eometric \textbf{G}rounding for autonomous \textbf{Driv}ing. As shown in Fig. \ref{fig-1}(c), VGGDrive employs a frozen 3D foundational model to map the multi-view image inputs inherent in autonomous driving tasks to 3D features with geometric consistency. To effectively bridge the 3D geometric features with the 2D visual representations in the LLM and enable the 3D features to deeply steer the VLM's performance in specific driving tasks, a plug-and-play Cross-view 3D Geometric Enabler (CVGE) is proposed. The CVGE decouples the base LLM architecture and introduces a hierarchical adaptive injection mechanism, which injects 3D features into the LLM’s 2D visual embeddings in a multi-level, structured manner, thereby establishing genuine geometric grounding within the model. As shown in Fig. \ref{fig-2}, extensive experiments demonstrate that the proposed VGGDrive achieves significant and consistent performance improvements across five mainstream autonomous driving language and trajectory evaluation benchmarks, covering tasks likes scene understanding, cross-view risk perception, motion prediction and trajectory planning.


\begin{itemize}
    \item We pioneer the integration of mature visual 3D foundation models into VLM-driven autonomous driving frameworks, effectively bridging the critical gap in cross-view geometric perception within this architecture.
    \item We propose a plug-and-play CVGE that facilitates deep coupling between 3D geometric features and VLMs through a hierarchical adaptive injection mechanism, establishing a solid geometric grounding for the model.
    \item Extensive experiments across five mainstream benchmarks thoroughly demonstrate the performance advantages of VGGDrive, affirming the feasibility and promising potential of this novel technical paradigm of empowering VLMs with 3D models for autonomous driving. 
\end{itemize}


\begin{figure*}[!t]
	\centering
	\includegraphics[width=6.8in]{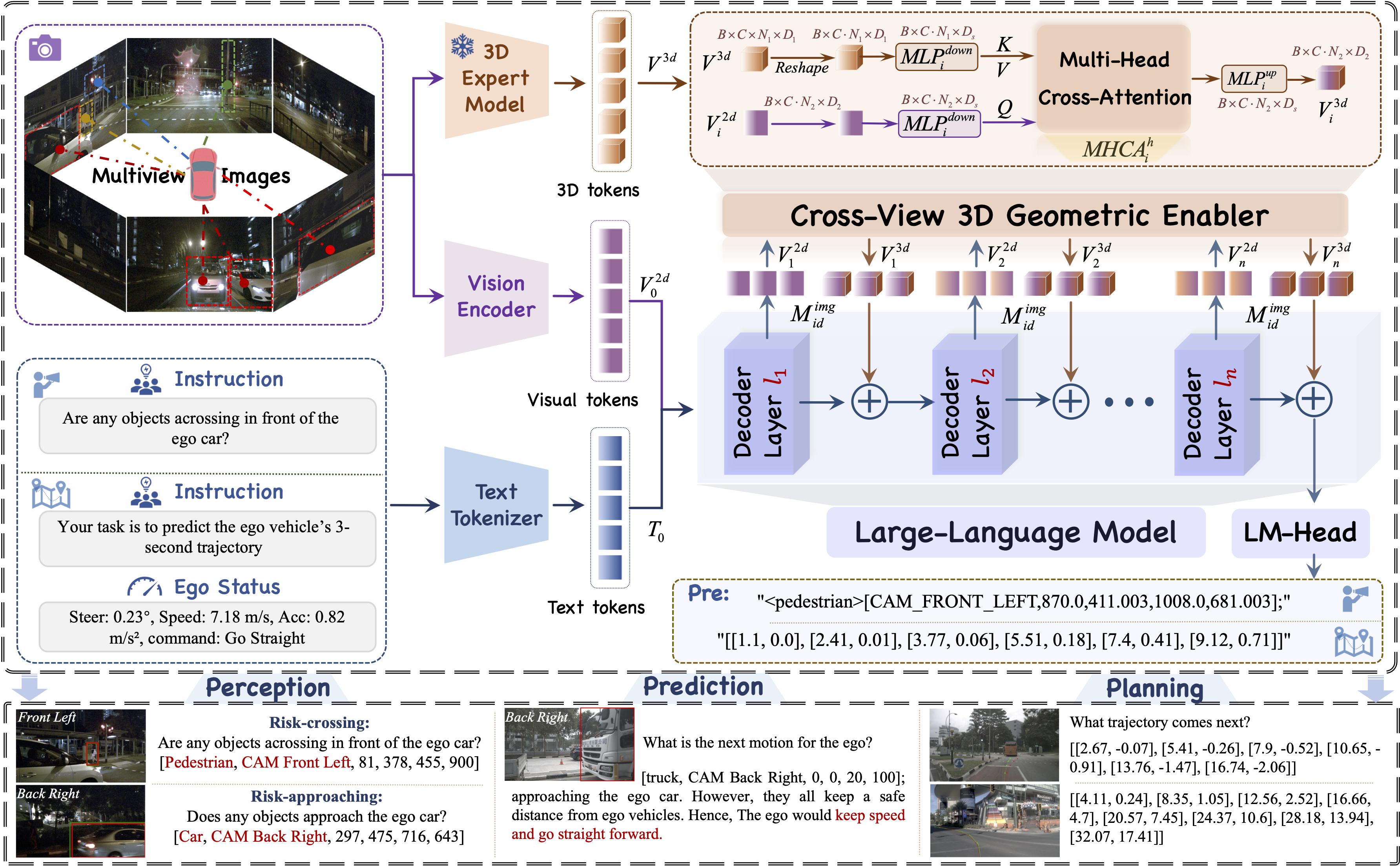}
	\vspace{-0.15in}
	\DeclareGraphicsExtensions.
	\begin{center}
		\caption{Overview of \textbf{VGGDrive}. Specifically, the frozen visual 3D foundation model (VGGT \cite{wang2025vggt}) extracts geometrically consistent 3D features $V^{3d}$ through cross-view analysis, while the base VLM is decomposed into multiple decoder layers. The proposed CVGE sequentially integrates the shared 3D features $V^{3d}$ with the 2D visual representations $V_{i}^{2d}$, injecting them $V_{i}^{3d}$ through a hierarchical adaptive mechanism, thereby establishing geometric grounding and enabling deep enhancement of the VLM architecture.} \label{fig-3}
	\end{center}
	\vspace{-0.35in}
\end{figure*}

\section{Related Work}
\label{sec:Related}

\subsection{VLMs for Autonomous Driving}
Vision-language models (VLMs) \cite{zhu2025internvl3, bai2025qwen2, wang2024qwen2} enhance the generalization and interpretability of traditional autonomous driving systems by leveraging rich world knowledge and powerful reasoning capabilities. Early studies \cite{jiang2024senna, jiang2025alphadrive, marcu2024lingoqa, ma2024dolphins} primarily employ VLMs as high-level scene understanding and decision-support tools, generating natural language descriptions of the environment or driving recommendations from visual inputs. With the maturation of end-to-end learning paradigms \cite{hu2023planning}, research has progressively evolved towards more promising vision-language-action (VLA) models \cite{renz2024carllava, fu2025orion, hwang2024emma, renz2025simlingo, huang2023fuller, xie2025vlms, hu2025vision}. CarLLaVA \cite{renz2024carllava} is a VLA model for autonomous driving that leverages camera inputs to predict both path and waypoint outputs.

However, VLMs struggle to improve performance in real-world autonomous driving tasks that require fine-grained cross-view spatial and geometric understanding. Some studies \cite{wang2025omnidrive, sima2024drivelm, ding2024holistic} teach VLMs spatial concepts using large-scale question-answer pairs, but this approach fails to equip solid geometric priors, resulting in limited improvements. Consequently, subsequent works \cite{li2025recogdrive, luo2025adathinkdrive} have resorted to adding a separate action decoder on top of VLMs to specifically handle trajectory prediction. While this improves trajectory performance, it disconnects scene understanding from action decision-making. In contrast to the two schemes, our VGGDrive aims to inject the cross-view spatial modeling capabilities of mature visual 3D foundation models into the base VLM, addressing its inherent limitations and adapting it for autonomous driving applications.

\subsection{Visual 3D Foundation Models}
Recently, a series of powerful visual 3D foundation models \cite{wang2025vggt, leroy2024grounding, wang2024dust3r} have emerged in the field of 3D vision, pre-trained on large-scale 3D datasets, demonstrating significant advantages in cross-view scene geometric modeling. VGGT \cite{wang2025vggt} is a feed-forward neural network that processes multi-view visual inputs to directly reconstruct 3D scenes, inferring key attributes such as camera parameters, depth maps, and point clouds. Given their powerful scene representation capabilities, some pioneering works \cite{huang2025mllms, zheng2025learning, abouzeid2025geoaware, li2025towards} have attempted to combine VLMs with these 3D foundation models, demonstrating significant potential in tasks such as indoor 3D scene understanding. 3DRS \cite{huang2025mllms} enhances VLM 3D capabilities by distilling 3D features from VGGT and aligning them with visual features from the final hidden states of the VLM. However, these methods \cite{huang2025mllms, zheng2025learning, li2025spatial, li2025towards} primarily focus on indoor, static, single-camera scenarios, which differ significantly from the dynamic, multi-camera environments of autonomous driving. Furthermore, their simplistic integration strategies fail to meet the accuracy and robustness required for complex driving tasks.

\vspace{-0.05in}
\section{Methodology}
To effectively inject and deeply empower the foundational VLM with the cross-view geometric grounding capability of VGGT, we propose the VGGDrive framework. As illustrated in Fig. \ref{fig-3}, the VGGDrive architecture consists of three core components: (1) Base VLM (Qwen2.5-VL), which processes both visual and textual inputs and generates the corresponding reasoning and action tokens through a unified autoregressive transformer decoder; (2) Hierarchical Adaptive Injection Mechanism, which decouples the structure of the base LLM, sequentially extracting its visual embeddings and feeding them into the CVGE, while adaptively injecting the 3D visual embeddings enhanced by CVGE into the corresponding hidden states; (3) Cross-view 3D Geometric Enabler, which is responsible for deeply integrating the 3D geometric features generated by VGGT with the 2D visual representations in the VLM, thereby enabling profound transfer of geometric knowledge. The key components of the model are detailed in the following sections.

\subsection{Base VLM Overview}
We adopt the Qwen2.5-VL-7B model \cite{bai2025qwen2} as the visual-language backbone for VGGDrive. Qwen2.5-VL is a series of powerful multimodal large language models, it not only demonstrates exceptional visual understanding capabilities but also benefits from its open-source nature, which facilitates fine-tuning for task-specific adaptations.

In this work, the VLM takes as input a set of N surround-view images $I = \{ {I_c}\} _{c = 1}^C$ and the corresponding language task instruction $L$. For benchmarks constructed from the nuScenes dataset (\textit{NuInstruct} \cite{ding2024holistic}, \textit{DriveLM} \cite{sima2024drivelm}, \textit{OmniDrive} \cite{wang2025omnidrive}, \textit{nuScenes-Plan} \cite{caesar2020nuscenes}), we use six surround-view images from the current frame ($C=6$). For trajectory prediction tasks based on the NAVSIM dataset \cite{dauner2024navsim}, we instead employ three front-view perspectives (namely front-left, front, and front-right, $C=3$). For the nuscenes-plan and NAVSIM trajectory planning tasks, additional ego-vehicle state information and navigation commands is incorporated into the textual instructions. During the feature extraction stage, multi-view images are processed by the visual encoder to produce 2D visual embeddings $V_{0}^{2d} \in {\mathbb{R}^{B \times C \cdot {N_2} \times {D_2}}}$, while the language instruction is encoded into text embeddings $T_0$ through the text encoder. These embeddings are concatenated (${x_0} = [V_0^{2d},{T_0}]$) and subsequently fed into a base LLM, which generates the corresponding textual tokens. The model is optimized by minimizing the standard cross-entropy loss:
\begin{equation}
\begin{array}{c}
{L_{CE}} =  - \sum\limits_{t = 1}^T {\log } \ {p_\theta }({y_t}|{y_{ < t}}, \ \{ {I_c}\} _{c = 1}^C, \ L),
\end{array}
\end{equation}
where ${y_t}$ is the t-th output token, and ${p_\theta }$ is the probability predicted by the model given all previous tokens and the multimodal context (i.e., images and instructions).

\subsection{Hierarchical Adaptive Injection Mechanism}
To fully harness the cross-view geometric modeling capability of VGGT and effectively empower the VLM to meet the stringent accuracy and robustness demands of complex autonomous driving scenarios, we design a Hierarchical Adaptive Injection Mechanism. Specifically, this mechanism employs the frozen VGGT model to perform cross-view 3D geometric modeling on an input set of N surround-view images $I = \{ {I_c}\} _{c = 1}^C$, from which the visual features extracted before the DPT module \cite{wang2025vggt} are utilized as the 3D features. Notably, we retain the original camera and registration embeddings within the 3D features $V^{3d} \in {\mathbb{R}^{B \times C \times {N_1} \times {D_1}}}$, as these embeddings also encode critical multi-view information that is indispensable for accurate scene geometry representation. Furthermore, our CVGE allows the 2D visual embeddings $V_{0}^{2d}$ to query the 3D representations $V^{3d}$, thereby capturing the necessary cross-view geometric information.

Subsequently, we decouple the architecture of the base LLM to ensure that the hidden states modeled at each decoder layer can be extracted as $X_{i}$.
\begin{equation}
\begin{array}{c}
{X_i} = D{L_{i}}({x_{i-1}}), \ \ i = 1,...,n,
\end{array}
\end{equation}
where $D{L_{i}}$ denotes the $i$-th decoder layer, and $n$ represents the total number of stacked decoder layers in the base LLM. The 2D visual representation $\{{ V_{i}^{2d} \in {\mathbb{R}^{B \times C \cdot {N_2} \times {D_2}}}}\}_{i = 1}^n$ from each layer is retrieved using fixed image ID positional masking $M_{id}^{img}$ applied to the hidden states $X_{i}$. 
\begin{equation}
\begin{array}{c}
V_i^{2d} = X{}_i, \ \ if \ M_{id}^{img} = 1 ,
\end{array}
\end{equation}
where $M_{id}^{img} \in {\{ 0,1\} ^{B \times (C \cdot {N_2} + N_s + N_t)}}$ denotes the image ID mask, with values set to 1 for image token $N_2$ positions and 0 otherwise (i.e., special tokens $N_s$ and text tokens $N_t$). Next, $V_{i}^{2d}$ and $V^{3d}$ are fed into the proposed CVGE to obtain geometry-enhanced 3D visual embeddings $\{{ V_{i}^{3d} \in {\mathbb{R}^{B \times C \cdot {N_2} \times {D_2}}}}\}_{i = 1}^n$. Considering the differences in embedding representations and their sensitivity to 3D information across network layers, the CVGE adopts a modular design with consistent structure but independent parameters across layers, enabling the 2D visual features at each layer to adaptively learn and extract the geometric information most relevant to that layer.
\begin{equation}
\begin{array}{c}
V_i^{3d} = CVG{E_i}({V^{3d}},V_i^{2d}), \ \ i = 1,...,n.
\end{array}
\end{equation}

Finally, with the assistance of the mask $M_{id}^{img}$ replaces $V_{i}^{2d}$ in the hidden states $X_{i}$, and the input to the next decoder layer $x_{i}$ is obtained through a residual connection.
\begin{equation}
\begin{array}{c}
{X_i^{'}} = \left\{ \begin{array}{l}
V_i^{3d}, \ \ if \ M_{id}^{img} = 1 \\
{X_i}, \ \  otherwise
\end{array} \right.,
\end{array}
\end{equation}
\begin{equation}
\begin{array}{c}
x_i = {X_i} + X_i^{'},  \ \ i = 1,...,n.
\end{array}
\end{equation}

\begin{table*}[htbp]
	\centering
	\normalsize
	\tabcolsep=0.28cm
	\renewcommand{\arraystretch}{0.98}
	\caption{The performance comparison on the NAVSIM \textit{navtest} \cite{dauner2024navsim} benchmark, evaluated using closed-loop metrics, involves both SOTA E2E approaches and existing VLA models under supervised fine-tuning. This evaluation aims to reflect the performance gains achieved by VGGDrive in enhancing the \textit{closed-loop trajectory planning} capability of the base VLM.} \label{tab-1}
	\vspace{-0.05in}
	\scalebox{0.9}{
    \begin{tabular}{lccccccccc}
        \toprule
        \textbf{NAVSIM} & Base Model & Image & \multicolumn{1}{c}{Lidar} & \multicolumn{1}{c}{NC↑} & \multicolumn{1}{c}{DAC↑} & \multicolumn{1}{c}{\textbf{EP↑}} & \multicolumn{1}{c}{TTC↑} & \multicolumn{1}{c}{Comf.↑} & \multicolumn{1}{c}{\textbf{PDMS↑}} \\
        \midrule
        TransFuser \cite{chitta2022transfuser} & \multirow{6}[2]{*}{E2E} & \multicolumn{1}{c}{$\surd$}     & \multicolumn{1}{c}{$\surd$} & 97.78 & 92.63 & 78.88 & 92.89 & 99.98 & 83.88 \\
        PARA-Drive \cite{weng2024drive} & \multicolumn{1}{c}{} & \multicolumn{1}{c}{$\surd$}     &       & 97.90 & 92.40 & 79.30 & 93.00 & 99.8  & 84.00 \\
        DRAMA \cite{yuan2024drama} & \multicolumn{1}{c}{} & \multicolumn{1}{c}{$\surd$}     & \multicolumn{1}{c}{$\surd$} & 98.19 & 95.18 & 81.33 & 94.17 & \textbf{100}   & 86.87 \\
        Hydra-MDP-V{$_{8192}$} \cite{li2024hydra} & \multicolumn{1}{c}{} & \multicolumn{1}{c}{$\surd$}     & \multicolumn{1}{c}{$\surd$} & 98.30 & 96.00 & 78.70 & 94.60 & \textbf{100}   & 86.50 \\
        DiffusionDrive \cite{liao2025diffusiondrive} & \multicolumn{1}{c}{} & \multicolumn{1}{c}{$\surd$}     & \multicolumn{1}{c}{$\surd$} & 98.20 & 96.20 & 82.20 & 94.70 & \textbf{100}   & 88.10 \\
        WoTE \cite{li2025end} & \multicolumn{1}{c}{} & \multicolumn{1}{c}{$\surd$}     & \multicolumn{1}{c}{$\surd$} & 98.50 & \textbf{96.80} & 81.90 & 94.90 & 99.9  & 88.30 \\
        \midrule
        Baseline & Qwen2.5-VL-7B & \multicolumn{1}{c}{$\surd$}     &       & 97.83 & 94.08 & 81.00 & 94.04 & 99.98 & 86.04 \\
        ImagiDrive \cite{li2025imagidrive} & LLaVA-1.6-7B & \multicolumn{1}{c}{$\surd$}     &       & 97.90 & 95.50 & 80.70 & 93.10 & 99.9  & 86.40 \\
        AutoVLA (SFT) \cite{zhou2025autovla} & Qwen2.5-VL-3B & \multicolumn{1}{c}{$\surd$}     &       & 96.89 & 94.43 & 75.82 & 88.06 & 99.94 & 80.54 \\
        ReCogDrive (SFT) \cite{li2025recogdrive} & InternVL3-8B & \multicolumn{1}{c}{$\surd$}     &       & 98.30 & 95.10 & 81.10 & 94.30 & \textbf{100}   & 86.80 \\
        AdaThinkDrive (SFT) \cite{luo2025adathinkdrive} & InternVL3-8B & \multicolumn{1}{c}{$\surd$}     &       & 98.50 & 94.40 & 79.90 & 94.90 & \textbf{100}   & 86.20 \\
        \midrule
        VGGT-Dist & Qwen2.5-VL-7B & \multicolumn{1}{c}{$\surd$}     &       & 97.84 & 94.81 & 81.30 & 94.42 & 99.98 & 86.68 \\
        VGGT-Add & Qwen2.5-VL-7B & \multicolumn{1}{c}{$\surd$}     &       & 97.81 & 94.07 & 80.84 & 94.40 & 99.98 & 86.10 \\
        \textbf{VGGDrive} & Qwen2.5-VL-7B & \multicolumn{1}{c}{$\surd$}     &       & \textbf{98.55} & 96.30 & \textbf{82.92} & \textbf{95.59} & 99.98 & \textbf{88.76} \\
        \bottomrule
		\end{tabular}%
	}
	\vspace{-0.02in}
\end{table*}%

\begin{table*}[ht!]
	\centering
	\normalsize
	\tabcolsep=0.042cm
	\renewcommand{\arraystretch}{0.98}
	\caption{The performance comparison on the NuInstruct dataset \cite{ding2024holistic} is conducted against existing SOTA methods. This experiment is crucial for evaluating the performance gains of VGGDrive in cross-view \textit{risk object perception} (\textbf{MAP}), \textit{state predictio}n, and \textit{ego-motion forecasting} within autonomous driving scenarios. The symbol * indicates $\max\left(\frac{\text{Accuracy} + \text{MAP} + \text{BLEU} - \text{MAE}}{4}, 0\right).$} \label{tab-2}
	\vspace{-0.05in}
	\scalebox{0.86}{
    \begin{tabular}{ccccccccccc}
        \toprule
        \multicolumn{1}{c}{\multirow{2}[4]{*}{Dataset}} & \multirow{2}[4]{*}{Metrics} & \multicolumn{4}{c}{Zero-shot Models} & \multicolumn{5}{c}{Fine-tuned Models} \\
        \cmidrule(lr){3-6} \cmidrule(lr){7-11}     & \multicolumn{1}{c}{} & \multicolumn{1}{c}{GPT-4o} \cite{hurst2024gpt} & \multicolumn{1}{c}{LLAVA-OV} \cite{li2024llava} & \multicolumn{1}{c}{RoboTron} \cite{huang2025robotron} & \multicolumn{1}{c}{Qwen2.5-VL} & \multicolumn{1}{c}{Baseline} & \multicolumn{1}{c}{InMLLM} \cite{ding2024holistic} & \multicolumn{1}{c}{VGGT-Dist} & \multicolumn{1}{c}{VGGT-Add} & \multicolumn{1}{c}{\textbf{VGGDrive}} \\
        \midrule
        \multicolumn{1}{c}{\multirow{5}[2]{*}{NuInstruct}} & \textbf{MAE}↓  & 9.93  & 87.04 & 19.36 & 24.10 & 4.35  & 9.08  & 3.73  & 3.63  & \textbf{3.08} \\
              & \textbf{Accuracy}↑ & 10.64 & 3.75  & 2.57 & 0.63  & 47.71 & 32.48 & 56.21 & 56.35 & \textbf{56.37} \\
              & \textbf{MAP↑} & 0     & 0     & 0  & 0   & 6.15  & 21.93 & 28.51 & 30.12 & \textbf{37.49} \\
              & BLEU↑ & 7.08  & 8.55  & 8.06  & 5.56 & 75.75 & 35.2  & 79.23 & 79.45 & \textbf{81.13} \\
              & \textbf{Average*↑} & 1.95  & 0     & 0  & 0   & 31.32 & 20.13 & 40.06 & 40.57 & \textbf{42.98} \\
        \bottomrule
		\end{tabular}%
	}
	\vspace{-0.02in}
\end{table*}%

\begin{table*}[ht!]
	\centering
	\normalsize
	\tabcolsep=0.038cm
	\renewcommand{\arraystretch}{0.98}
	\caption{The performance comparison on the DriveLM dataset \cite{sima2024drivelm} is conducted against existing SOTA methods. This experiment is crucial for evaluating the performance gains of VGGDrive in cross-view risk object perception (\textbf{Match}), action prediction and planning.} \label{tab-3}
	\vspace{-0.05in}
	\scalebox{0.86}{
    \begin{tabular}{cccccccccccc}
    \toprule
        \multicolumn{1}{c}{\multirow{2}[4]{*}{Dataset}} & \multirow{2}[4]{*}{Metrics} & \multicolumn{3}{c}{Zero-shot Models} & \multicolumn{7}{c}{Fine-tuned Models} \\
        \cmidrule(lr){3-5} \cmidrule(lr){6-12}          & \multicolumn{1}{c}{} & \multicolumn{1}{c}{GPT-4o} & \multicolumn{1}{c}{LLAVA-OV} & \multicolumn{1}{c}{RoboTron} & \multicolumn{1}{c}{Baseline} & \multicolumn{1}{c}{DriveLM} & \multicolumn{1}{c}{TM-LMM} \cite{ishaq2025tracking} & \multicolumn{1}{c}{OminiDrive} & \multicolumn{1}{c}{FSDrive} \cite{zeng2025FSDdrive} & \multicolumn{1}{c}{I-VL4Drive} \cite{li2024driving} & \multicolumn{1}{c}{VGGDrive} \\
        \midrule
        \multicolumn{1}{c}{\multirow{5}[2]{*}{DriveLM}} & \textbf{Accuracy}↑ & 38.55 & 25.03 & 52.94 & 64.35 & 52.30 & 59.60 & 70.00 & 71.77 & 73.39 & \textbf{77.50} \\
          & ChatGPT↑ & \textbf{67.27} & 65.70 & 51.11 & 65.30 & 55.35 & 58.44 & 65.00 & 63.42 & 65.25 & 64.76 \\
          & Language↑ & 8.97  & 14.44 & 7.12  & 44.90 & 45.13 & 46.38 & \multicolumn{1}{p{4.19em}}{-} & 52.77 & 48.56 & \textbf{56.69} \\
          & \textbf{Match}↑ & 24.00 & 40.93 & 35.05 & 34.54 & 35.73 & 35.73 & 37.00 & 39.19 & 47.65 & \textbf{49.77} \\
          & \textbf{Average↑} & 41.21 & 42.36 & 39.47 & 54.59 & 47.20 & 51.50 & 56.00 & 57.12 & 60.02 & \textbf{61.26} \\
    \bottomrule
		\end{tabular}%
	}
	\vspace{-0.05in}
\end{table*}%

\subsection{Cross-view 3D Geometric Enabler}
Existing integration schemes \cite{huang2025mllms, zheng2025learning} based on simple feature concatenation or addition fail to fully enable VLM visual embeddings to capture the rich scene geometry embedded within VGGT, limiting their adaptability in complex and highly dynamic autonomous driving scenarios. To address this, we design the CVGE, which establishes a learnable cross-modal interaction mechanism. This enables 2D visual embeddings to autonomously excavate and integrate critical information from 3D geometric features, thereby achieving deep geometric empowerment of visual representations. Building on the previous discussion, the CVGE takes 2D visual embeddings $\{{ V_{i}^{2d} \in {\mathbb{R}^{B \times C \cdot {N_2} \times {D_2}}}}\}_{i = 1}^n$ and shared 3D geometric features $V^{3d} \in {\mathbb{R}^{B \times C \times {N_1} \times {D_1}}}$ as inputs, and outputs the geometry-enhanced 3D visual embeddings ${ V_{i}^{3d} \in {\mathbb{R}^{B \times C \cdot {N_2} \times {D_2}}}}$.

\begin{table*}[htbp]
	\centering
	\normalsize
	\tabcolsep=0.115cm
	\renewcommand{\arraystretch}{0.98}
	\caption{The performance comparison on the OmniDrive dataset \cite{wang2025omnidrive} focuses on caption-related tasks in which base VLMs excel.} \label{tab-4}
	\vspace{-0.06in}
	\scalebox{0.86}{
    \begin{tabular}{ccccccccccc}
        \toprule
        \multicolumn{1}{c}{\multirow{2}[4]{*}{Dataset}} & \multirow{2}[4]{*}{Metrics} & \multicolumn{3}{c}{Zero-shot Models} & \multicolumn{6}{c}{Fine-tuned Models} \\
        \cmidrule(lr){3-5} \cmidrule(lr){6-11}          & \multicolumn{1}{c}{} & \multicolumn{1}{c}{GPT-4o} & \multicolumn{1}{c}{LLAVA-OV} & \multicolumn{1}{c}{RoboTron} & \multicolumn{1}{c}{OmniDrive} & HERMES \cite{zhou2025hermes} & \multicolumn{1}{c}{Baseline} & \multicolumn{1}{c}{VGGT-Dist} & \multicolumn{1}{c}{VGGT-Add} & \multicolumn{1}{c}{VGGDrive} \\
        \midrule
        \multicolumn{1}{r}{\multirow{4}[2]{*}{OmniDrive}} & BLEU↑ & 10.91 & 16.14 & 20.30 & \textbf{38.00} & -     & 37.29 & 37.28 & 37.23 & 37.58 \\
              & CIDEr↑ & 24.42 & 28.41 & 34.33 & 68.60 & \multicolumn{1}{c}{74.10} & 86.29 & 86.41 & 86.38 & \textbf{86.57} \\
              & ROUGE↑ & 22.34 & 22.14 & 23.67 & 32.60 & \multicolumn{1}{c}{32.70} & 34.33 & 34.32 & 34.26 & \textbf{34.40} \\
              & \textbf{Average↑} & 19.22 & 22.23 & 26.10 & 46.40 & -     & 52.64 & 52.67 & 52.62 & \textbf{52.85} \\
        \bottomrule
		\end{tabular}%
	}
	\vspace{-0.025in}
\end{table*}%

\begin{table*}[htbp]
	\centering
	\normalsize
	\tabcolsep=0.3cm
	\renewcommand{\arraystretch}{0.95}
	\caption{Performance comparison on nuScenes open-loop planning, with metrics from BEV-Planner's reproduced results \cite{li2024ego}.} \label{tab-5}
	\vspace{-0.06in}
	\scalebox{0.86}{
    \begin{tabular}{lccccccccccccc}
        \toprule
        \multicolumn{2}{c}{\multirow{1}[4]{*}{\textbf{nuScenes}}} & \multicolumn{4}{c}{L2} & \multicolumn{4}{c}{Collision (\%) ↓} & \multicolumn{4}{c}{Intersection (\%) ↓} \\
        \cmidrule(lr){3-6} \cmidrule(lr){7-10} \cmidrule(lr){11-14}    \multicolumn{2}{c}{} & \multicolumn{1}{c}{1s} & \multicolumn{1}{c}{2s} & \multicolumn{1}{c}{3s} & \multicolumn{1}{c}{avg} & \multicolumn{1}{c}{1s} & \multicolumn{1}{c}{2s} & \multicolumn{1}{c}{3s} & \multicolumn{1}{p{3em}}{avg} & \multicolumn{1}{c}{1s} & \multicolumn{1}{c}{2s} & \multicolumn{1}{c}{3s} & \multicolumn{1}{c}{avg} \\
        \midrule
        UniAD \cite{hu2023planning} & \multicolumn{1}{c}{\multirow{3}[2]{*}{E2E}} & 0.20  & 0.42  & 0.75  & 0.46  & 0.02  & 0.25  & 0.84  & 0.37  & 0.20  & 1.33  & 3.24  & 1.59 \\
        VAD-Base \cite{jiang2023vad} &       & 0.17  & 0.34  & 0.60  & 0.37  & 0.04  & 0.27  & 0.67  & 0.33  & 0.21  & 2.13  & 5.06  & 2.47 \\
        BEV-Planner \cite{li2024ego} &       & 0.16  & 0.32  & 0.57  & 0.35  & 0.00  & 0.29  & 0.73  & 0.34  & 0.35  & 2.62  & 6.51  & 3.16 \\
        \midrule
        Baseline \cite{li2024ego} & \multicolumn{1}{c}{\multirow{6}[2]{*}{VLA}} & \textbf{0.14} & 0.34  & 0.60  & 0.36  & 0.06  & 0.21  & 0.96  & 0.41  & 0.66  & 2.58  & 5.43  & 2.89 \\
        OmniDrive-Q \cite{wang2025omnidrive} &       & \textbf{0.14} & 0.29  & 0.55  & 0.33  & \textbf{0.00} & 0.13  & 0.78  & 0.30  & 0.56  & 2.48  & 5.96  & 3.00 \\
        OmniDrive-L \cite{wang2025omnidrive} &       & 0.15  & 0.36  & 0.70  & 0.40  & 0.06  & 0.27  & 0.72  & 0.35  & \textbf{0.49} & \textbf{1.99} & 4.86  & 2.45 \\
        VGGT-Dist &       & \textbf{0.14} & 0.30  & 0.56  & 0.33  & 0.02  & 0.20  & 0.80  & 0.34  & 0.64  & 2.46  & 5.68  & 2.93 \\
        VGGT-Add &       & \textbf{0.14} & 0.30  & 0.55  & 0.33  & 0.02  & 0.18  & 0.74  & 0.31  & 0.68  & 2.42  & 5.61  & 2.90 \\
        VGGDrive  &       & \textbf{0.14} & \textbf{0.28} & \textbf{0.51} & \textbf{0.31} & 0.02  & \textbf{0.10} & \textbf{0.55} & \textbf{0.22} & 0.63  & 2.27  & \textbf{4.02} & \textbf{2.31} \\
        \bottomrule
		\end{tabular}%
	}
	\vspace{-0.1in}
\end{table*}%

First, the shared $V^{3d}$ are flattened so that tokens from all views are aligned to match the total number of tokens in the 2D visual embeddings $V_{i}^{2d}$, facilitating cross-view information interaction. Subsequently, to optimize computational efficiency and reconcile the dimensional differences between the two embeddings, we employ two independent MLPs for dimensionality reduction. The reduced $V_{i}^{2d}$ features are used as query vectors $Q$, while the reconstructed and reduced $V^{3d}$ features serve as key $K \in {\mathbb{R}^{B \times C \cdot {N_1} \times {D_s}}}$ and value $V \in {\mathbb{R}^{B \times C \cdot {N_1} \times {D_s}}}$ vectors, respectively.
\begin{equation}
\begin{array}{c}
Q = MLP_i^{down}(V_i^{2d})\in {\mathbb{R}^{B \times C \cdot {N_2} \times {D_s}}},
\end{array}
\end{equation}
\begin{equation}
\begin{array}{c}
K,V =  MLP_i^{down}(Re({V^{3d}}))\in {\mathbb{R}^{B \times C \cdot {N_2} \times {D_s}}},
\end{array}
\end{equation}
where $D_s$ denotes the reduced dimension (i.e., ${{\rm{D}}_s} = {D_1}/s$), and $s$ represents the scaling factor.

It is noteworthy that in autonomous driving tasks, camera intrinsic and extrinsic parameters are typically treated as known prior information. These parameters are critical for trajectory planning tasks that rely on complete 3D scene mapping, such as NAVSIM \cite{dauner2024navsim} and NuScenes-Plan \cite{caesar2020nuscenes}. Therefore, we explicitly encode the camera parameters and incorporate them into the generated key ($K$) and value ($V$) vectors. Specifically, the camera intrinsic parameters are first scaled and calibrated based on the target image dimensions. Then, the extrinsic parameters are combined to construct a homogeneous transformation matrix. Finally, $T_i^{img2lidar} \in {\mathbb{R}^{B \times C \times 4 \times 4}}$ represents the geometric transformation from the image coordinate system to the LiDAR coordinate system, which is obtained by computing the inverse of the matrix.
\begin{equation}
\begin{array}{c}
T_i^{img2lidar} = {\left( {{K_i} \cdot \left[ {\begin{array}{*{20}{c}}
{R_i^T}&{ - R_i^T{t_i}}\\
0&1
\end{array}} \right]} \right)^{ - 1}},
\end{array}
\end{equation}
where $K_i$ is the intrinsic matrix of the $i$-th camera, $R_i$ is the rotation matrix from the LiDAR coordinate system to the $i$-th camera coordinate system, and $t_i$ is the translation vector from the LiDAR to the $i$-th camera coordinate system. The transformation matrix $T_i^{img2lidar}$ is flattened to form camera tokens, which is then encoded by an MLP to generate the ground truth camera embedding $Cam_{gt} \in {\mathbb{R}^{B \times C \times {D_s}}}$. $Cam_{gt}$ is added to the first token of the corresponding viewpoint in the previously generated $K$ and $V$ (i.e., the camera token estimated by VGGT \cite{wang2025vggt}) to form the new $K$ and $V$.

Based on the obtained $Q$, $K$, and $V$ representations, we design a cross-modal geometric attention fusion module. Unlike traditional static fusion methods such as simple feature addition or concatenation, the multi-head attention mechanism allows the model to autonomously uncover long-range, deep correlations between 2D visual features and 3D geometric representations, while performing on-demand information fusion through dynamic weight computation. This design bridges the semantic gap between 2D and 3D modalities effectively through the query-key semantic matching mechanism, thereby facilitating a shift from ``passive reception" to ``active exploration." Consequently, 2D visual features ($Q$) can autonomously extract the most relevant spatial information from 3D geometric representations ($K$ and $V$), thus enabling true deep empowerment.

The fused features are then passed through an MLP ($MLP_i^{up}$) to increase the dimensionality, ensuring that the final dimensions align with $V_i^{2d}$, thus obtaining the 3D visual representations ${ V_{i}^{3d} \in {\mathbb{R}^{B \times C \cdot {N_2} \times {D_2}}}}$ empowered by cross-view geometric information.
\begin{equation}
\begin{array}{c}
V_i^{3d} = MLP_i^{up}(MHCA_i^h(Q,K,V)),
\end{array}
\end{equation}
where $MHCA_i^h$ denotes the multi-head cross-attention mechanism that fuses cross-modal information at the $i$-th layer, with $h$ representing the number of attention heads, typically set to 8. The scale factor $s$ in $MLP_i^{up}$ is set to 4.

\begin{figure*}[!t]
	\centering
	\includegraphics[width=6.83in]{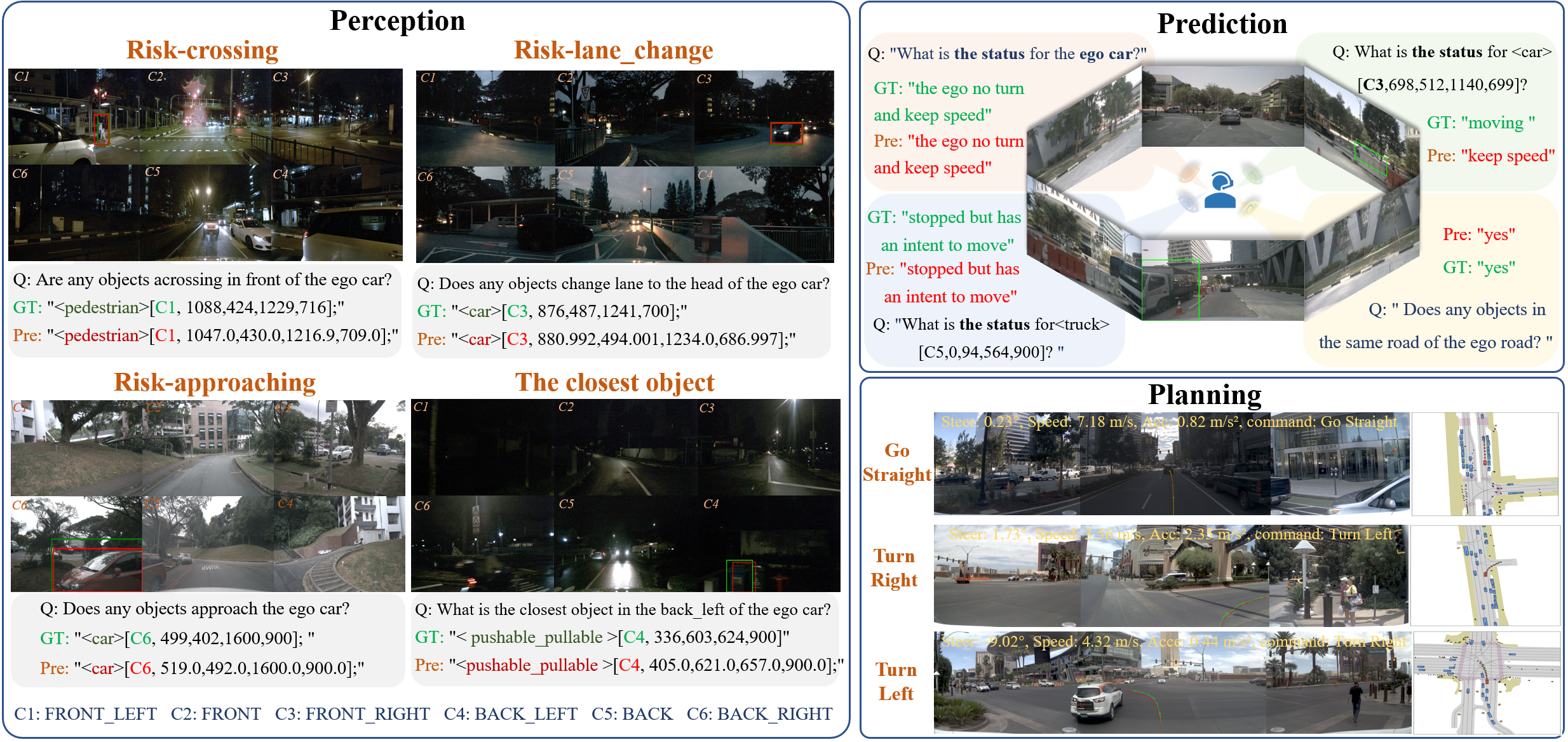}
	\vspace{-0.15in}
	\DeclareGraphicsExtensions.
	\begin{center}
		\caption{Visualization of VGGDrive's performance across various autonomous driving attribute evaluation tasks.} \label{fig-4}
	\end{center}
	\vspace{-0.4in}
\end{figure*}
\section{Experiment}
\subsection{Implementation Details}
We choose Qwen2.5-VL-7B \cite{bai2025qwen2} as the base model. The training process consists of two stages, with the parameters of the VGGT \cite{wang2025vggt} frozen throughout both stages. In the first stage, we freeze the base VLM parameters and train only the parameters introduced by CVGE for 2 epochs, using a learning rate of $1 \times {10^{ - 4}}$ and a batch size of 2. In the second stage, we fine-tune both the VLM and CVGE parameters for 2 epochs with a learning rate of $5 \times {10^{ - 5}}$ , while keeping the batch size the same. The dimensionality reduction scaling factor $s$ in CVGE is set to 4. All experiments are conducted using 16 NVIDIA H200 GPUs.

For both training and inference, we use the surrounding view of the current frame. Specifically, for the NuScenes evaluation benchmark, we utilize a 6-view surrounding image, and for NAVSIM, we use a 3-view front-facing image. For open-loop trajectory planning tasks in NuScenes-Plan \cite{caesar2020nuscenes} and NAVSIM \cite{dauner2024navsim}, we include the ego-vehicle state (e.g., velocity and acceleration) and the command (e.g., Go Straight, Turn Left, Turn Right) information in the prompt.

\subsection{Dataset and Metric}
To comprehensively evaluate the performance of VGGDrive in enhancing the VLM with cross-view 3D geometric capabilities across various autonomous driving tasks, we conduct experiments on five mainstream autonomous driving benchmarks. These benchmarks include language evaluation benchmarks: NuInstruct \cite{ding2024holistic}, DriveLM \cite{sima2024drivelm}, and Omnidrive \cite{wang2025omnidrive}, as well as two trajectory planning tasks: NuScenes-Plan \cite{caesar2020nuscenes} and NAVSIM \cite{dauner2024navsim}. The NuInstruct and DriveLM datasets contain numerous cases for evaluating cross-view risk target perception, risk target state prediction, and ego motion prediction (e.g., Fig. \ref{fig-4}). The Omnidrive dataset primarily focuses on scene description tasks related to captioning in driving scenarios. The NuScenes-Plan and NAVSIM datasets focus on open-loop and closed-loop trajectory planning tasks, respectively. To ensure a fair comparison and comprehensively reflect the advantages of VGGDrive, we train and evaluate its performance on five different datasets using the corresponding metrics.

\subsection{Performance Comparison}
\textbf{NNAVSIM.} As reported in Table 1, the closed-loop trajectory comparison is shown. 1) Compared to the base VLM, VGGDrive achieves a \textbf{2.72} improvement in PDMS, while VGGT-Dist and VGGT-Add show only marginal performance gains. 2) Existing VLA methods typically fine-tune with large-scale driving Q$\&$A data during the first stage and resume with trajectory training in the second stage. Compared to these methods after two-stage SFT, VGGDrive, which requires training only on the trajectory task, still achieves a performance gain of nearly 2. 3) E2E methods inherently have an advantage in trajectory planning tasks, yet VGGDrive still performs comparably. Without exaggeration, VGGDrive achieves the optimal performance for VLM-based autoregressive trajectory generation with a PDMS of 88.76. These findings indicate that improving trajectory performance by empowering the base VLM with cross-view geometric capability is a viable path, rather than relying solely on the addition of an auxiliary action decoder.

\textbf{NuInstruct.} From Tab. \ref{tab-2}, it is evident that the base VLM struggles with tasks like cross-view risk object perception (MAP) and state prediction in autonomous driving, with only limited improvement after fine-tuning on specific datasets. Notably, the impressive performance of VGGT-Dist and VGGT-Add on this dataset highlights the effectiveness and promising potential of VGGT features for autonomous driving tasks. Compared to the base VLM, VGGDrive achieves a \textbf{31.34} improvement in the critical MAP metric, surpassing SOTA methods by \textbf{7.37}.

\textbf{DriveLM.} This benchmark is crucial for evaluating a model's capability in cross-view object perception, action prediction and planning. As reported in Table \ref{tab-3}, VGGDrive substantially surpasses the baseline, improving the Match and Average metrics by \textbf{15.23} and \textbf{6.67}, respectively, and also exceeds current SOTA methods by 2.12 and 1.24.

\textbf{OmniDrive.} This dataset primarily focuses on caption-related tasks in autonomous driving scenarios, where the base VLM, after fine-tuning on the specific dataset, can achieve impressive performance. As shown in Tab. \ref{tab-4}, VGGDrive, while incorporating cross-view capabilities, does not compromise the base model's ability in caption-related tasks, despite the difficulty of simultaneously excelling in both aspects, as indicated in \cite{wang2025omnidrive}.

\textbf{NuScenes.} It primarily reflects the model's capability in open-loop trajectory planning. As shown in Tab. \ref{tab-5}, VGGDrive outperforms existing SOTA methods, particularly achieving an 8$\%$ improvement in collision rate.

VGGDrive’s consistent advantages across multiple benchmarks indicate that leveraging 3D foundation models to empower VLM-based autonomous driving tasks is both meaningful and holds significant potential for the future.

\begin{table}[t]
	\centering
	\normalsize
	\tabcolsep=0.075cm
	\renewcommand{\arraystretch}{1.0}
	\caption{Ablation study of various integration schemes between VGGT and VLM on the NAVSIM \cite{dauner2024navsim} and NuInstruct Datasets \cite{ding2024holistic}. Inference speed is evaluated on the NuInstruct \textit{MAP} task.} \label{tab-6}
	\vspace{-0.08in}
	\scalebox{0.84}{
    \begin{tabular}{clcccccc}
        \toprule
        \multicolumn{1}{c}{\multirow{2}[4]{*}{ID}} & \multirow{2}[4]{*}{Methods} & \multicolumn{2}{c}{NAVSIM} & \multicolumn{4}{c}{NuInstruct} \\
        \cmidrule(lr){3-4} \cmidrule(lr){5-7} \cmidrule(lr){8-8}         & \multicolumn{1}{c}{} & \multicolumn{1}{c}{EP↑} & \multicolumn{1}{c}{PDMS↑} & \multicolumn{1}{c}{MAP↑} & \multicolumn{1}{c}{BLEU↑} & \multicolumn{1}{c}{Avg↑} & \multicolumn{1}{c}{Time (s)↓} \\
        \midrule
        1     & Baseline & 81.00 & 86.04 & 6.15  & 75.75 & 31.32 & \textbf{0.81} \\
        2     & VGGT & 79.56 & 84.45 & 3.34 & 74.69 & 30.35 & 0.84 \\
        3     & VGGT-Dist & 81.30 & 86.68 & 28.51 & 79.23 & 40.06 & 0.82 \\
        4     & VGGT-Add & 80.84 & 86.10 & 30.12 & 79.45 & 40.57 & 0.87 \\
        5     & VGGT-MHCA & 81.94 & 87.83 & 32.35 & 79.39 & 40.85 & 0.91 \\
        6     & OURS  & \textbf{82.92} & \textbf{88.76} & \textbf{37.49} & \textbf{81.13} & \textbf{42.98} & 1.04 \\
        \bottomrule
		\end{tabular}%
	}
	\vspace{-0.25in}
\end{table}%

\subsection{Ablation Studies}
Extensive ablation experiments are conducted on the NAVSIM \cite{dauner2024navsim} and NuInstruct \cite{ding2024holistic} datasets to validate the effectiveness and robustness of our integration approach and the key models within VGGDrive.

\textbf{Effectiveness of the Integration Scheme.} Tab. \ref{tab-6} presents the ablation study of the integration scheme between VGGT and VLM. Specifically, \textit{ID-2} refers to directly replacing the visual encoder with VGGT's 3D features; \textit{ID-3} \cite{huang2025mllms, li2025spatial} refers to aligning the visual features in the hidden states of the LLM output with 3D features using a distillation approach; \textit{ID-4} \cite{zheng2025learning} refers to adding the visual-encoded features and 3D features together before feeding them into the LLM; \textit{ID-5} \cite{lin2025evo} refers to using multi-head cross-attention to fuse the 2D visual features and 3D features before passing them to the LLM. The performance results across both datasets demonstrate that our approach more effectively empowers the VLM to handle complex and highly dynamic autonomous driving scenarios.

\begin{table}[t]
	\centering
	\normalsize
	\tabcolsep=0.077cm
	\renewcommand{\arraystretch}{1.0}
	\caption{Ablation Study of the Main Components of VGGDrive.} \label{tab-7}
	\vspace{-0.08in}
	\scalebox{0.84}{
    \begin{tabular}{clcccccc}
        \toprule
        \multicolumn{1}{c}{\multirow{2}[4]{*}{ID}} & \multirow{2}[4]{*}{Methods} & \multicolumn{2}{c}{NAVSIM} & \multicolumn{4}{c}{NuInstruct} \\
        \cmidrule(lr){3-4} \cmidrule(lr){5-7} \cmidrule(lr){8-8}         & \multicolumn{1}{c}{} & \multicolumn{1}{c}{EP↑} & \multicolumn{1}{c}{PDMS↑} & \multicolumn{1}{c}{MAP↑} & \multicolumn{1}{c}{BLEU↑} & \multicolumn{1}{c}{Avg↑} & \multicolumn{1}{c}{Time (s)↓} \\
        \midrule
        1     & Baseline & 81.00 & 86.04 & 6.15  & 75.75 & 31.32 & \textbf{0.81} \\
        2     & w/o MHCA & 82.05 & 87.79 & 27.43 & 77.89 & 39.29 & 1.00 \\
        3     & Share CVGE & 82.48 & 88.05 & 30.85 & 79.55 & 40.95 & 0.96 \\
        4     & w/o Residual & 82.32 & 87.86 & 27.98 & 78.31 & 40.02 & 1.03 \\
        5     & One-stage SFT & 82.83 & 88.62 & 36.19 & 80.75 & 42.05 & 1.04 \\
        6     & OURS  & \textbf{82.92} & \textbf{88.76} & \textbf{37.49} & \textbf{81.13} & \textbf{42.98} & 1.04 \\
        \bottomrule
		\end{tabular}%
	}
	\vspace{-0.2in}
\end{table}%

\textbf{Effectiveness of Key Components.} Tab. \ref{tab-7} presents the ablation analysis of the key components in VGGDrive. Under the integration scheme and adaptive injection mechanism of VGGDrive, \textit{ID-2} refers to the ablation of MHCA, where addition is used as a replacement for MHCA; \textit{ID-3} refers to the scenario where the multi-layer CVGE shares its structure and parameters. \textit{ID-4} corresponds to removing the residual setting during LLM injection (i.e., Eq. 6: $\{{x_i = X_i^{'}}\}_{i = 1}^n$), and \textit{ID-5} corresponds to using only single-stage full fine-tuning of all model parameters.

\begin{table}[htbp]
	\centering
	\normalsize
	\tabcolsep=0.18cm
	\renewcommand{\arraystretch}{1.0}
    \vspace{-0.08in}
	\caption{Ablation Study of the 3D Expert Model.} \label{tab-8}
	\vspace{-0.08in}
	\scalebox{0.84}{
        \begin{tabular}{ccccccc}
        \toprule
        \textbf{NAVSIM} & \multicolumn{1}{c}{NC↑} & \multicolumn{1}{c}{DAC↑} & \multicolumn{1}{c}{EP↑} & \multicolumn{1}{c}{TTC↑} & \multicolumn{1}{c}{Comf.↑} & \multicolumn{1}{c}{PDMS↑} \\
        \midrule
        Baseline & 97.83 & 94.08 & 81.00 & 94.04 & 99.98 & 86.04 \\
        Fast3r \cite{yang2025fast3r} & 98.34 & 95.83 & 82.62 & 95.58 & 99.98 & 88.36 \\
        VGGT \cite{wang2025vggt}  & 98.55 & 96.30 & 82.92 & 95.59 & 99.98 & \textbf{88.76} \\
        \bottomrule
		\end{tabular}%
	}
	\vspace{-0.1in}
\end{table}%

\textbf{Effectiveness of the 3D Expert Model.} We also perform an ablation study of the 3D expert model, replacing VGGT with Fast3r, a 3D expert model developed concurrently with VGGT but exhibiting slightly inferior performance. Table \ref{tab-8} shows that incorporating Fast3r also leads to noticeable performance improvements, although it slightly underperforms VGGT across autonomous driving tasks. This further suggests the significant potential of incorporating 3D models to empower VLMs in addressing autonomous driving tasks, with the performance of stronger 3D expert models expected to yield even better results.

\begin{table}[htbp]
	\centering
	\normalsize
	\tabcolsep=0.14cm
	\renewcommand{\arraystretch}{1.0}
    \vspace{-0.08in}
	\caption{Further Ablation Analysis on the Navsim Benchmark.} \label{tab-9}
	\vspace{-0.08in}
	\scalebox{0.84}{
        \begin{tabular}{ccccccc}
        \toprule
        \textbf{NAVSIM} & \multicolumn{1}{c}{NC↑} & \multicolumn{1}{c}{DAC↑} & \multicolumn{1}{c}{EP↑} & \multicolumn{1}{c}{TTC↑} & \multicolumn{1}{c}{Comf.↑} & \multicolumn{1}{c}{PDMS↑} \\
        \midrule
        Baseline & 97.83 & 94.08 & 81.00 & 94.04 & 99.98 & 86.04 \\
        w/o $T_i^{img2lidar}$ & 98.33 & 95.93 & 82.59 & 95.42 & 99.99 & 88.32 \\
        s=8   & 98.17 & 96.15 & 82.66 & 95.26 & 99.99 & 88.36 \\
        s=2   & 98.50 & 96.16 & 82.80 & 95.78 & 99.99 & 88.70 \\
        OURS  & 98.55 & 96.30 & 82.92 & 95.59 & 99.98 & \textbf{88.76} \\
        \bottomrule
		\end{tabular}%
	}
	\vspace{-0.1in}
\end{table}%

Tab. \ref{tab-9} presents the ablation analysis on incorporating camera parameters $T_i^{img2lidar}$ in the trajectory task and the hyperparameter study of the scale factor $s$ (default set to 4) in CVGE. Further analysis of VGGDrive's integration strategy, injection mechanism, and core components demonstrates their effective collaboration in meeting diverse autonomous driving evaluation requirements.

\vspace{-0.02in}
\section{Conclusion}
In this paper, we propose the innovative architecture \textbf{VGGDrive}: Empowering \textbf{V}ision-Language Models with Cross-View \textbf{G}eometric \textbf{G}rounding for Autonomous \textbf{Driv}ing. The core of this architecture lies in the design of the \textbf{CVGE}, which effectively integrates cross-modal features and employs a hierarchical adaptive injection mechanism to deeply empower VLMs. Extensive experiments across five mainstream autonomous driving benchmarks demonstrate that VGGDrive achieves comprehensive and significant performance improvements across various autonomous driving attribute evaluation protocols. Most importantly, VGGDrive validates the feasibility and potential of empowering VLMs with cross-view geometric capabilities using 3D foundation models for autonomous driving. Compared to methods like constructing large-scale Q$\&$A datasets to teach VLMs or introducing independent action decoders, VGGDrive pioneers a distinct technical pathway for deploying VLMs in the autonomous driving domain.


{
    \small
    \bibliographystyle{ieeenat_fullname}
    \bibliography{main}
}

\clearpage
\setcounter{page}{1}
\maketitlesupplementary

\renewcommand{\thetable}{S1}
\begin{table*}[bp]
	\centering
	\normalsize
	\tabcolsep=0.3cm
	\renewcommand{\arraystretch}{1.0}
    \vspace{-0.05in}
	\caption{Training and Testing Sample Statistics of Five Mainstream Autonomous Driving Datasets Used by VGGDrive, and the Model Capabilities Evaluated on Each Dataset.} \label{tab-a1}
	\scalebox{0.98}{
        \begin{tabular}{ccccccc}
        \toprule
        Dataset & \multicolumn{1}{c}{\parbox{1.5cm}{\centering Train \\ Samples}} & \multicolumn{1}{c}{\parbox{1.5cm}{\centering Test \\ Samples}} & \multicolumn{1}{c}{\parbox{1.8cm}{\centering Cross-view \\ perception}} & \multicolumn{1}{c}{\parbox{2.1cm}{\centering Action\&States \\ prediction}} & \multicolumn{1}{c}{\parbox{1.8cm}{\centering Trajectory \\ planning}} & \multicolumn{1}{c}{\parbox{2cm}{\centering Scene \\ understanding}} \\
        \midrule
        NAVSIM \cite{dauner2024navsim} & 103.3k & 12.1k &       &       & \checkmark     &  \\
        NuInstruct \cite{ding2024holistic} & 71.8k & 16.1k & \checkmark     & \checkmark     &       &  \\
        DriveLM \cite{sima2024drivelm} & 376.2k & 15.5k & \checkmark     & \checkmark     &       & \checkmark \\
        OmniDrive \cite{wang2025omnidrive} & 318.4k & 54.1k &       &       &       & \checkmark \\
        nuScenes \cite{li2024ego} & 28.1k & 6.0k  &       &       & \checkmark     &  \\
        \bottomrule
		\end{tabular}%
	}
	\vspace{-0.05in}
\end{table*}

To further supplement the main content of this paper, we have organized additional material to elaborate on some key details. Specifically, Sec. \ref{sec-1} provides further clarification on the application of VGGDrive across five mainstream autonomous driving benchmarks and the corresponding evaluation metrics. Sec. \ref{sec-2} presents an interesting ablation analysis on the NAVSIM closed-loop trajectory planning task, where 3D features $V^{3d}$ are injected into different decoding layers $\{{ DL_{i} }\}_{i = 1}^n$ of the VLM. Sec. \ref{sec-3} presents additional visual samples that highlight the model's performance on critical tasks in autonomous driving. Tab.~\ref{tab-a1} presents the training and testing sample statistics of five mainstream autonomous driving datasets used by VGGDrive, along with the model capabilities evaluated on each dataset.

\renewcommand{\thesection}{A}
\section{Dataset and Metric}
\label{sec-1}
To assess the performance of VGGDrive across various attributes within the autonomous driving domain, we conduct evaluations on five prominent benchmarks. These benchmarks cover task-specific scenarios, including scene understanding, cross-view risk object perception, action and state prediction, and trajectory planning. In the tables within the main text, we highlight key metrics related to the model's cross-view 3D capabilities by using bold formatting.

\textbf{NAVSIM.} This dataset \cite{dauner2024navsim} is a real-world, planning-focused dataset derived from OpenScene, which is a compact version of nuPlan, the largest publicly available annotated driving dataset (1,192 and 136 scenarios for training and testing). NAVSIM is designed to evaluate the performance of autonomous driving systems in complex, dynamic scenarios. It utilizes a combination of eight cameras providing a 360° field of view (FOV) and a merged LiDAR point cloud from five sensors.  NAVSIM specifically targets challenging driving situations with dynamic driving intentions, while deliberately excluding simpler, static scenarios such as stationary scenes or constant-speed driving. The NAVSIM benchmark provides a nonreactive simulation environment and employs the Predictive Driver Model Score (PDMS) as its closed-loop planning metric:
\begin{equation}
\begin{array}{c}
PDMS = NC \times DAC \times \left( {\frac{{5 \times EP + 5 \times TTC + 2 \times C}}{{12}}} \right)
\end{array}
\end{equation}
where PDMS integrates five sub-metrics: No At-Fault Collision (NC), Drivable Area Compliance (DAC), Time-to Collision (TTC), Comfort (C), and Ego Progress (EP) to produce a comprehensive closed-loop planning score.

\renewcommand{\thefigure}{S1}
\begin{figure*}[!t]
    \centering
    \includegraphics[width=0.33\textwidth]{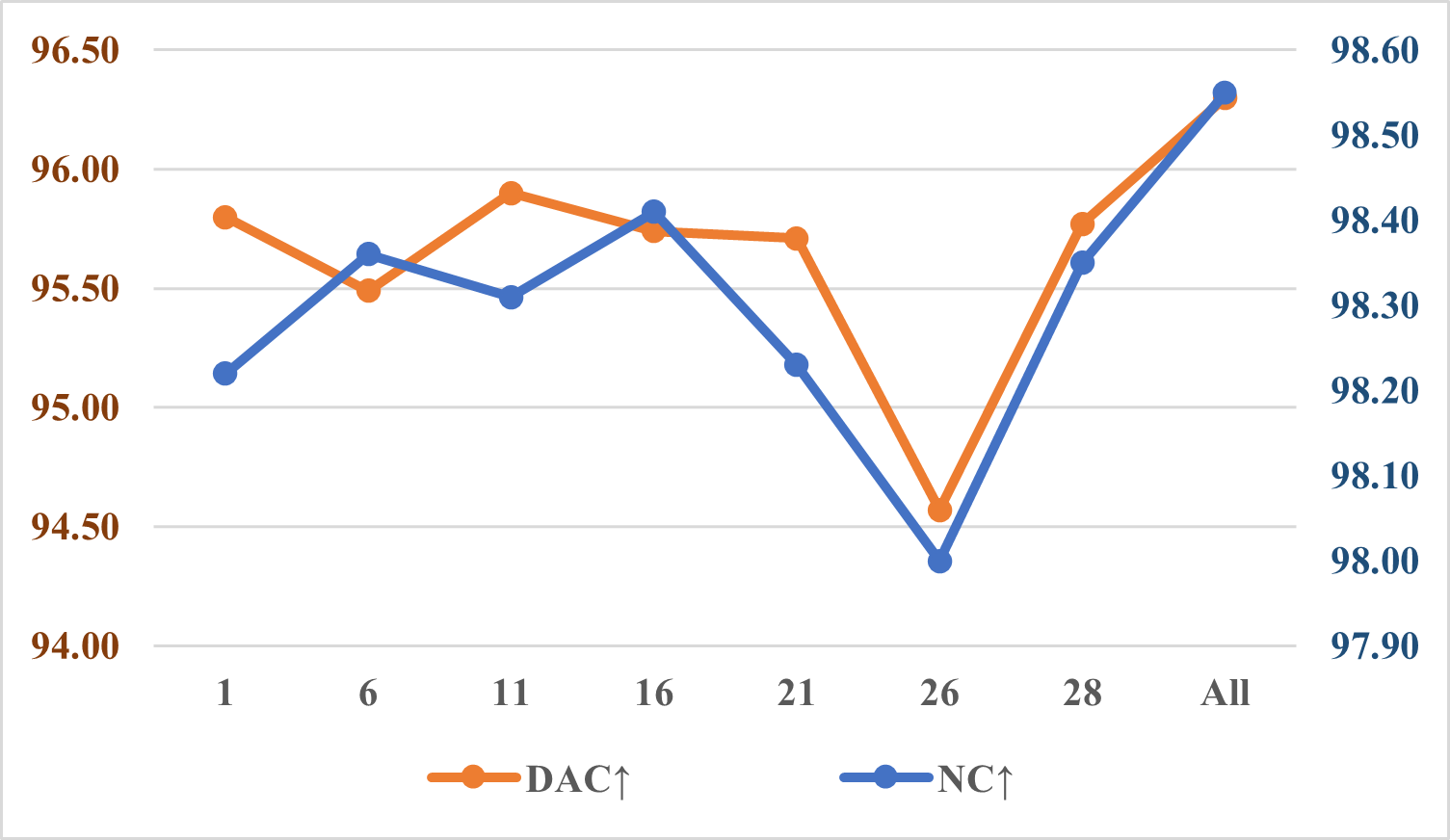}
    \includegraphics[width=0.33\textwidth]{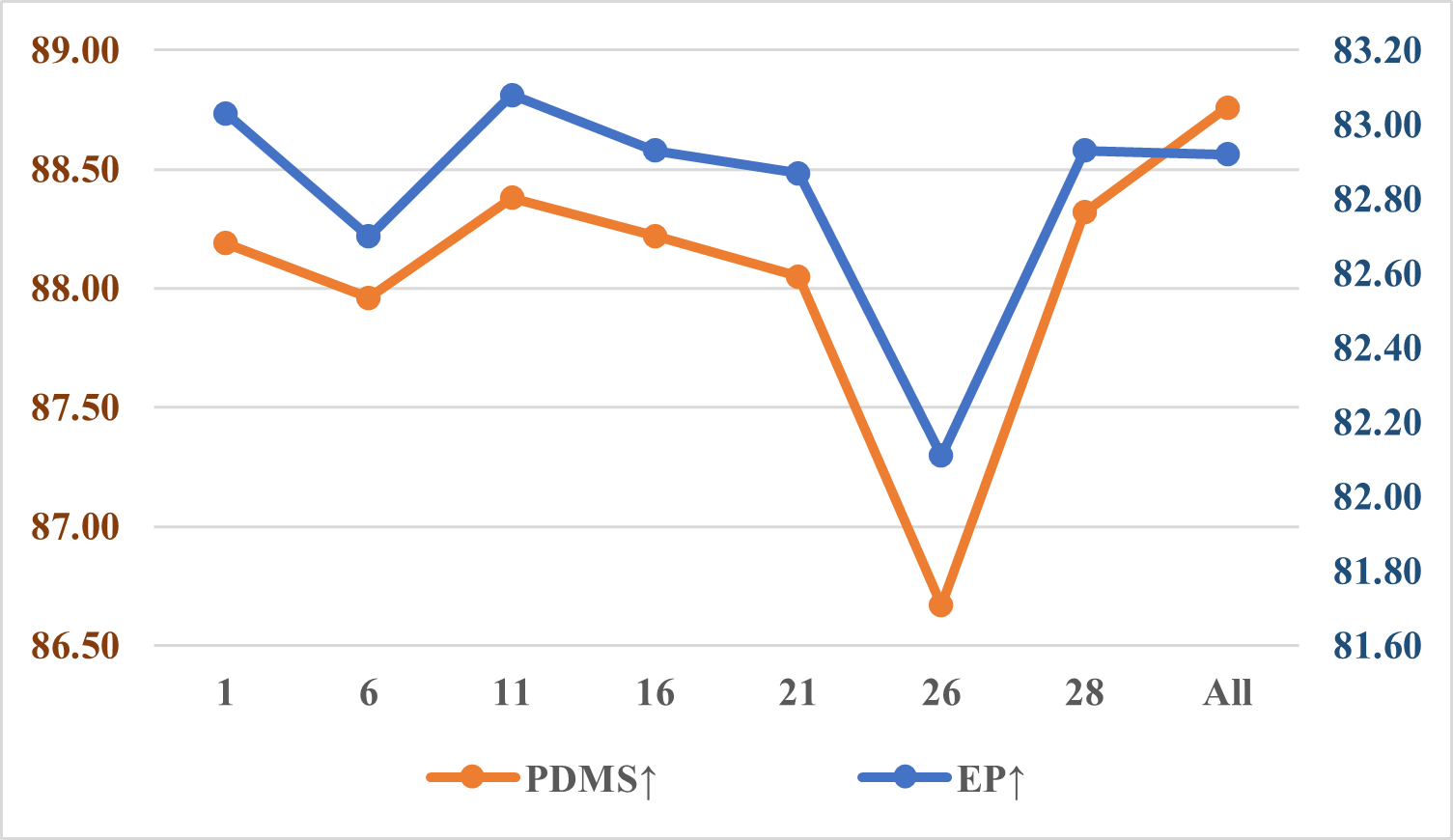}
    \includegraphics[width=0.33\textwidth]{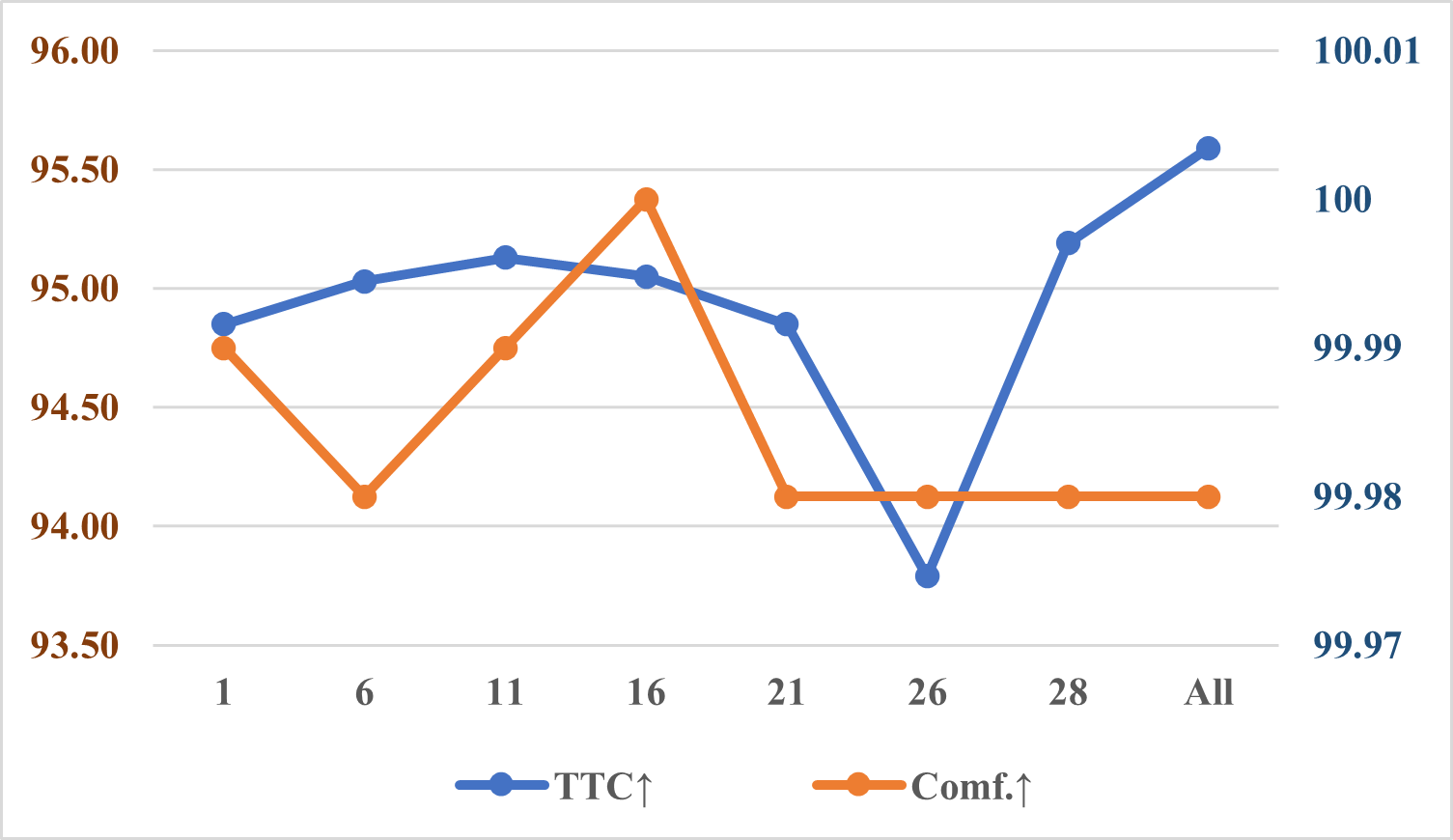}
    \vspace{-0.1in}
    \caption{Ablation analysis of closed-loop trajectory planning performance on the NAVSIM dataset when cross-view 3D geometric empowerment and adaptive injection are applied to individual decoding layers of the LLM.}
    \label{fig-s1}
    \vspace{-0.15in}
\end{figure*}

\textbf{NuInstruct.} This dataset \cite{ding2024holistic} samples a total of 11,850 keyframes from 850 videos within the NuScenes dataset. It includes a wide range of challenging samples, such as cross-view risk perception, target distance estimation, agent and ego state prediction, target motion prediction, and reasoning tasks. The dataset effectively reflects the model's ability to perform cross-view understanding and analyze 3D geometric scene information. NuInstruct uses a variety of metrics to evaluate different tasks, including Mean Absolute Error (MAE) for regression tasks, accuracy for classification tasks, Mean Average Precision (mAP) for detection tasks, and a rule-based BLEU metric for captioning tasks.

\textbf{DriveLM.} This dataset \cite{sima2024drivelm} is built upon the real-world driving dataset nuScenes and covers interactive scenarios involving vehicles, pedestrians, and traffic infrastructure on urban roads. The keyframes focus on moments that mark changes in driving intentions, such as acceleration, deceleration, and turning. The data samples not only include ``what objects are currently present" (cross-view perception), but also encompass ``how objects will move in the future" (action $\&$ states prediction), ``what the vehicle should do" (planning), ``specific behavior classifications" (e.g., fast driving straight, slow right turn), and ``trajectory coordinates" (motion). This dataset similarly reflects the model's ability to perform cross-view perception and analyze 3D geometric scene understanding. DriveLM implements four evaluation metrics: accuracy, LLM score, language rule-based evaluation, and match score.

\textbf{OmniDrive.} This dataset \cite{wang2025omnidrive} is also built upon the nuScenes dataset and primarily focuses on scene description, attention-based question answering, counterfactual reasoning question answering, decision planning question answering, and general dialogue. Due to the limitations in the availability of evaluation data and metrics, we selected scene description and general dialogue from this dataset to train and evaluate the model's ability to understand autonomous driving scene captions. The dataset primarily includes three language evaluation metrics: BLEU, CIDEr, and ROUGE, with the ``average" representing the mean of the three metrics.

\textbf{nuScenes.} We conduct open-loop trajectory planning experiments on the challenging public nuScenes dataset \cite{li2024ego}, which contains 1,000 driving scenes, each lasting approximately 20 seconds. The scene images are captured by six cameras, providing a 360° horizontal field of view (FOV), with keyframes annotated at a frequency of 2Hz. We follow the standard training and testing setup, adhering to the evaluation metrics used in OmniDrive. Among these metrics \cite{wang2025omnidrive, li2024ego}, L2 and Collision Rate are widely used, and we additionally introduce the Intersection Rate, which calculates the rate of collisions or intersections between the predicted trajectories and curbs (road boundaries).

\renewcommand{\thesection}{B}
\section{VLM Layer Injection Ablation}
\label{sec-2}

To validate the effectiveness of our hierarchical adaptive injection mechanism and explore the performance of individual decoding layers when injected with 3D features, we conduct further ablation experiments. Specifically, while maintaining identical configurations as the final VGGDrive model, we perform cross-view 3D geometric feature empowerment and adaptive injection on individual decoding layers $\{{ DL_{i} }\}_{i = 1}^n$. For the Qwen2.5-VL-7B model, which contains 28 decoding layers (i.e., $n=28$), we select different layers at regular intervals and observe the performance of closed-loop trajectory planning on the NAVSIM dataset.

1) Comparison of Fig. \ref{fig-s1} and Tab. \ref{tab-1} (in the main text): Using our cross-view 3D Geometric Enabler and decoupled adaptive injection mechanism, a significant performance improvement is observed when injecting features into a single layer (PDMS around 88), compared to the base VLM model (PDMS = 86.04). Meanwhile, this method also outperforms existing VGGT and VLM integration solutions (PDMS $<$ 87). Compared to single-layer injection, our VGGDrive achieves even better performance through full-layer adaptive injection. These results highlight the positive impact of VGGT features for multi-view tasks in autonomous driving and validate the effectiveness and robustness of our proposed VGGDrive design. 

2) Analysis of performance across different layers: By analyzing the performance variations across different layers, we observe significant differences in closed-loop planning performance when injecting 3D features into different decoding layers. Overall, a peak performance is achieved around layer 11, with reasonable performance maintained at both ends. The observed variations across layers also offer valuable insights for future research, suggesting that a reasonable trade-off between efficiency and performance could be achieved by leveraging just a single layer.

\renewcommand{\thesection}{C}
\section{Qualitative Results}
\label{sec-3}

This section further presents several visual examples to demonstrate VGGDrive’s capability in trajectory planning and action or state prediction within complex autonomous driving scenarios. Fig. \ref{fig-s2}, \ref{fig-s3}, and \ref{fig-s4} show trajectory planning visualizations for left turns, right turns, and straight driving, respectively, in the NAVSIM benchmark. Fig. \ref{fig-s5} provides additional examples of VGGDrive's open-loop trajectory performance in the nuScenes benchmark. In Fig. \ref{fig-s6}, we further illustrate how VGGDrive predicts the states and actions of both the ego vehicle and surrounding agents after perceiving and modeling cross-view scenes. These visualizations effectively highlight VGGDrive's advantage in injecting VGGT’s cross-view 3D scene features into VLM-driven autonomous driving tasks, showcasing substantial performance improvements. This validates the effectiveness and promising prospects of this novel technical approach, which avoids the reliance on constructing large-scale VQA datasets or adding additional trajectory-generation action decoders.

\renewcommand{\thefigure}{S2}
\begin{figure*}[!t]
	\centering
	\includegraphics[width=6.85in]{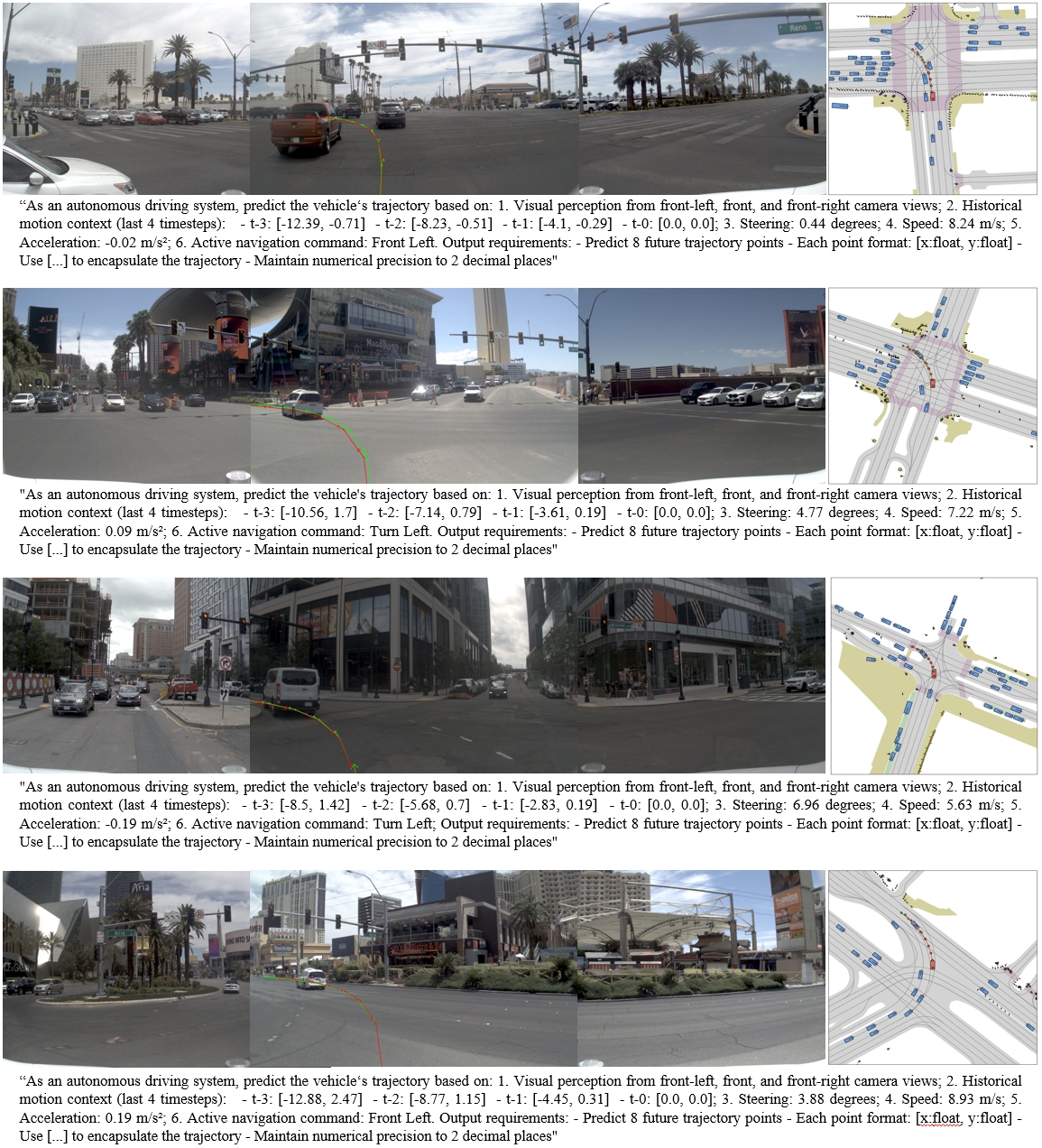}
    \vspace{-0.3in}
	\DeclareGraphicsExtensions.
	\begin{center}
		\caption{Qualitative results on the closed-loop trajectory planning task in the Navtest benchmark are presented, showcasing a typical left-turn example to demonstrate the performance of VGGDrive in complex driving scenarios.} \label{fig-s2}
	\end{center}
\end{figure*}

\renewcommand{\thefigure}{S3}
\begin{figure*}[!t]
	\centering
	\includegraphics[width=6.85in]{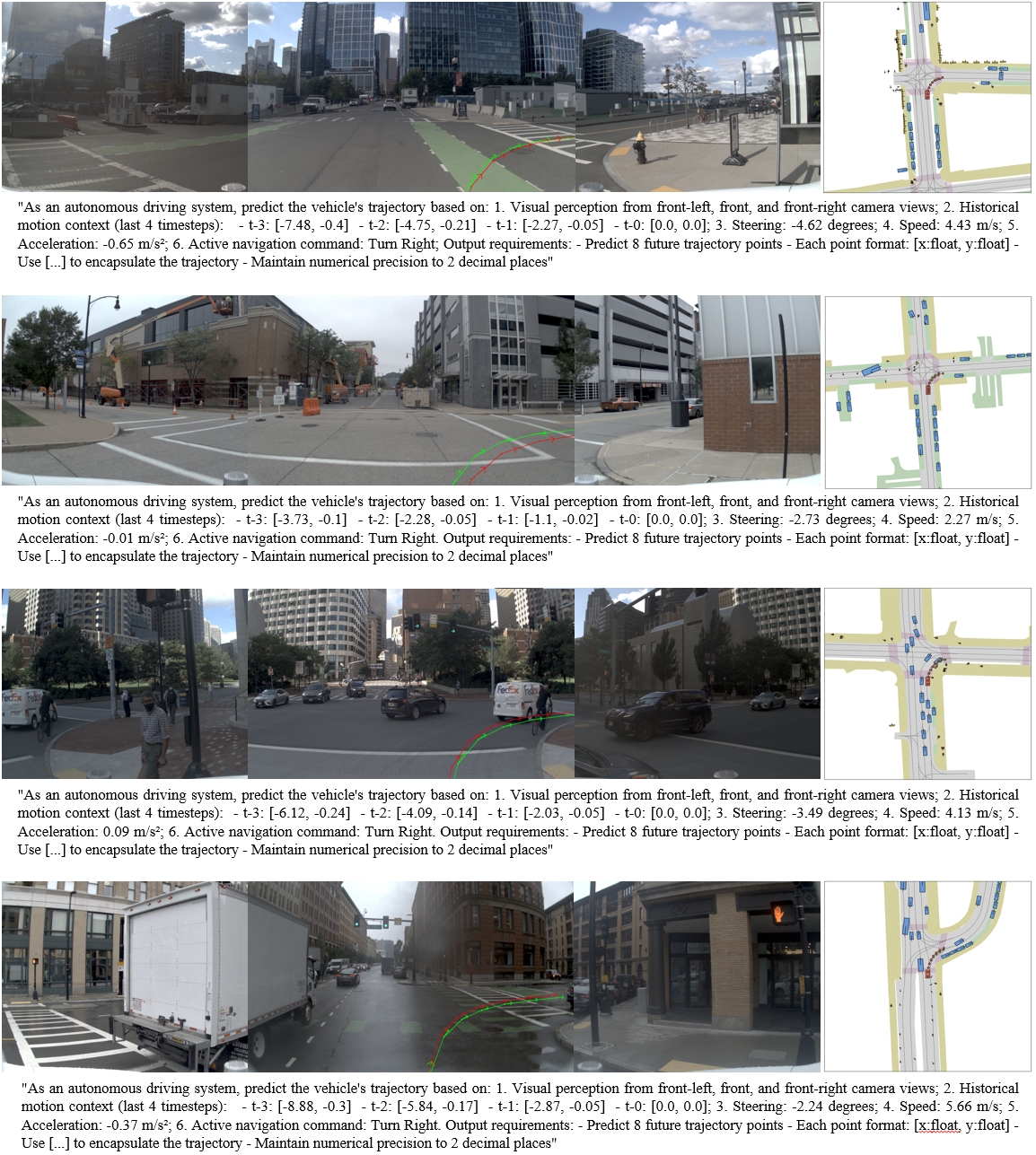}
    \vspace{-0.3in}
	\DeclareGraphicsExtensions.
	\begin{center}
		\caption{Qualitative results on the closed-loop trajectory planning task in the Navtest benchmark are presented, showcasing a typical right-turn example to demonstrate the performance of VGGDrive in complex driving scenarios.} \label{fig-s3}
	\end{center}
\end{figure*}

\renewcommand{\thefigure}{S4}
\begin{figure*}[!t]
	\centering
	\includegraphics[width=6.85in]{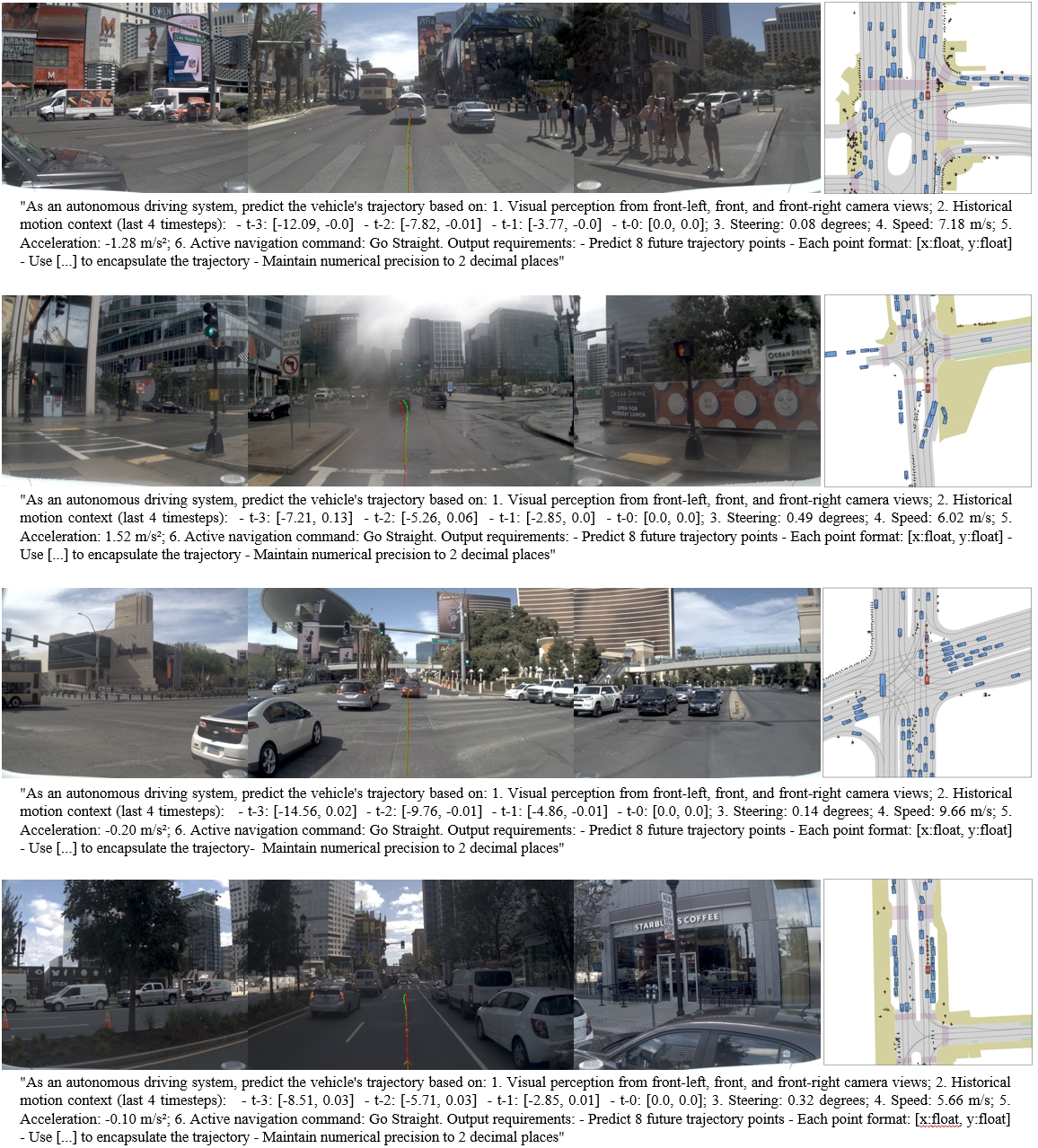}
    \vspace{-0.3in}
	\DeclareGraphicsExtensions.
	\begin{center}
		\caption{Qualitative results on the closed-loop trajectory planning task in the Navtest benchmark are presented, showcasing a typical straight-ahead example to demonstrate the performance of VGGDrive in complex driving scenarios.} \label{fig-s4}
	\end{center}
\end{figure*}

\renewcommand{\thefigure}{S5}
\begin{figure*}[htbp]
	\centering
	\includegraphics[width=6.85in]{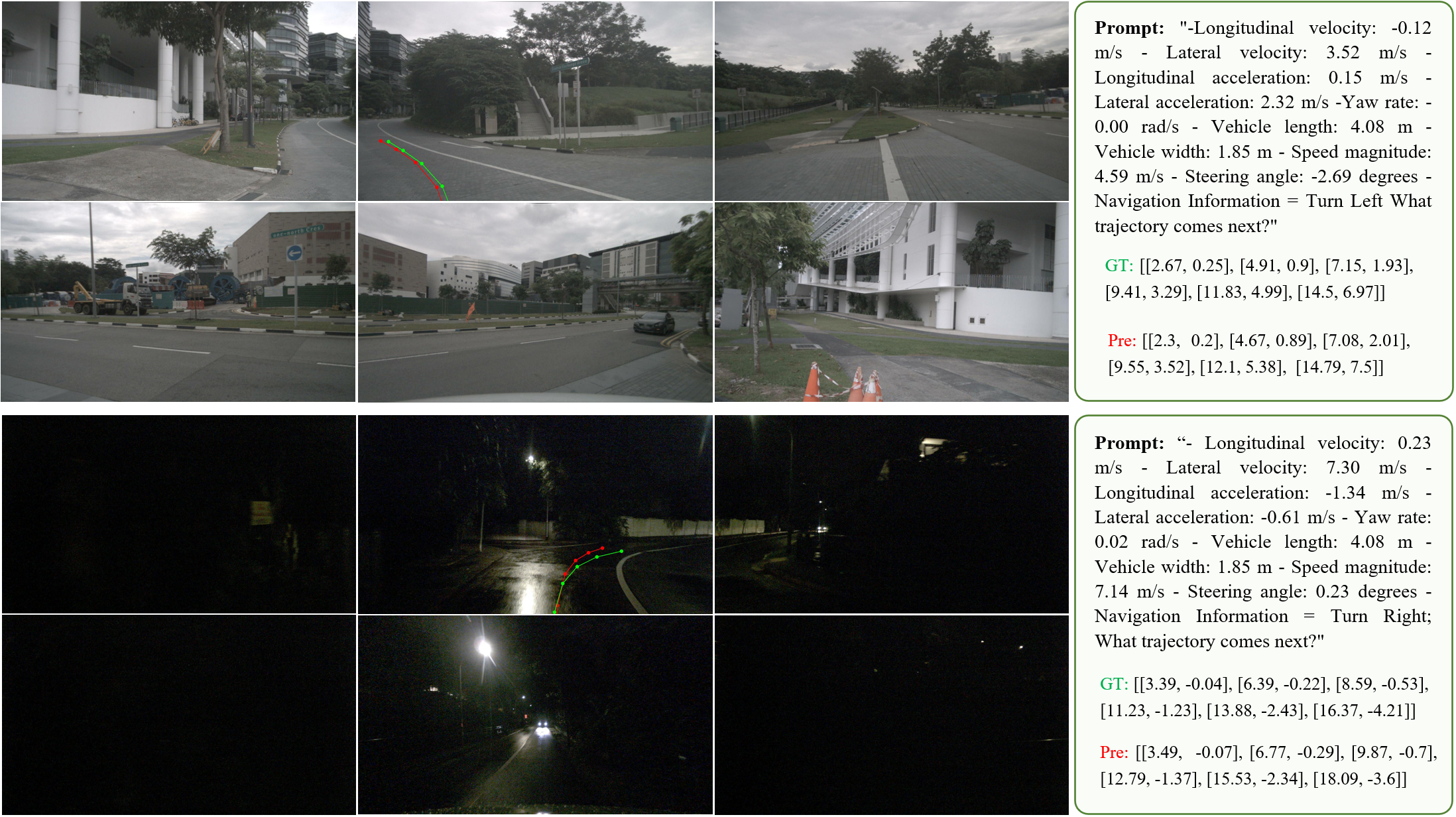}
    \vspace{-0.3in}
	\DeclareGraphicsExtensions.
	\begin{center}
		\caption{Qualitative results on the open-loop trajectory planning task in the nuScenes benchmark are presented.} \label{fig-s5}
	\end{center}
    \vspace{-0.3in}
\end{figure*}

\renewcommand{\thefigure}{S6}
\begin{figure*}[htbp]
	\centering
	\includegraphics[width=6.85in]{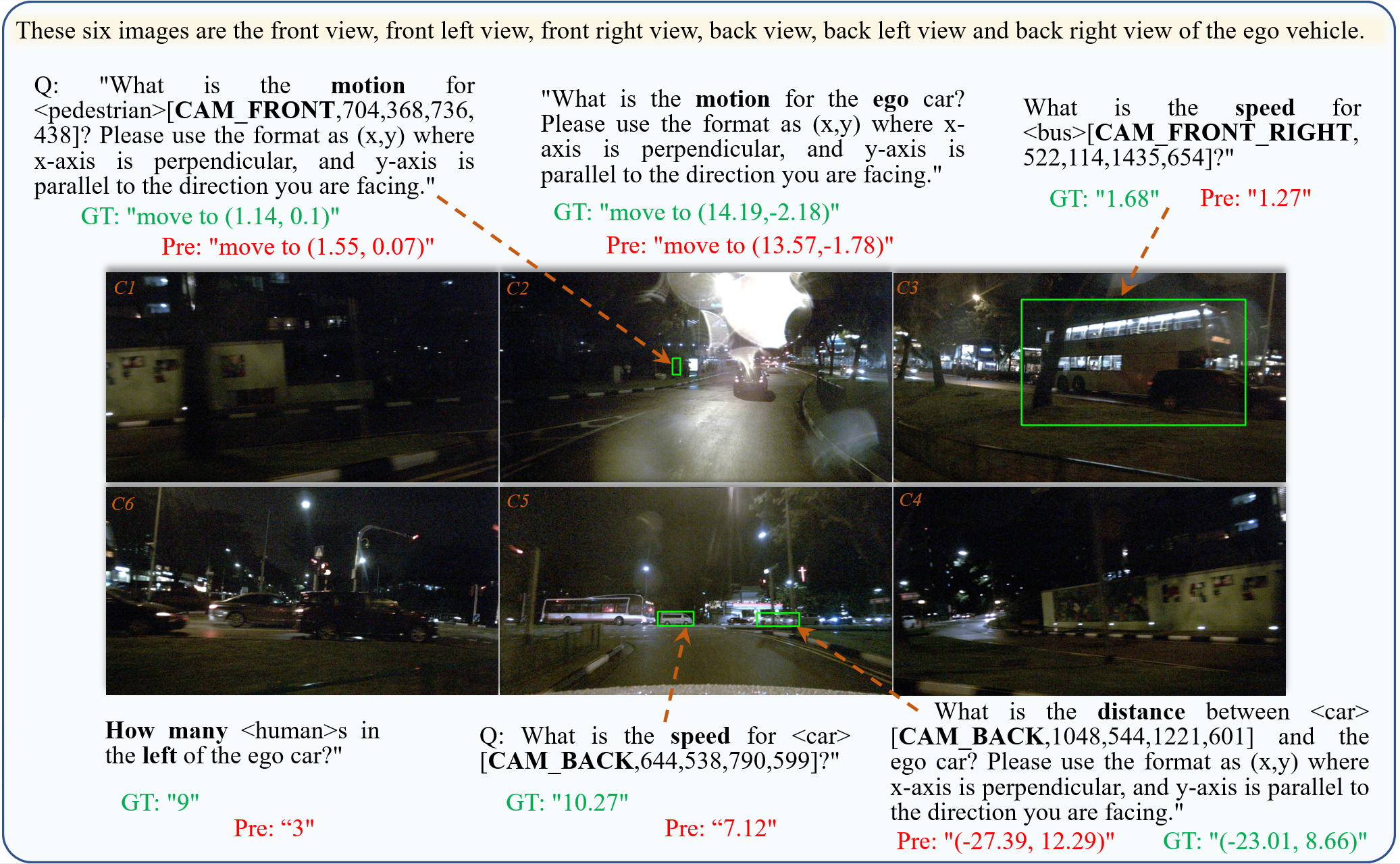}
    \vspace{-0.3in}
	\DeclareGraphicsExtensions.
	\begin{center}
		\caption{Qualitative results on Action $\&$ States prediction in autonomous driving scenarios are presented.} \label{fig-s6}
	\end{center}
    \vspace{-0.3in}
\end{figure*}

\end{document}